\documentclass{article}

\PassOptionsToPackage{numbers, compress}{natbib}
\usepackage[preprint]{neurips_2026}

\usepackage[utf8]{inputenc}  
\usepackage[T1]{fontenc}     
\usepackage{hyperref}        
\usepackage{url}             
\usepackage{booktabs}        
\usepackage{amsmath,amssymb,amsthm,mathrsfs}
\usepackage{amsfonts}        
\usepackage{nicefrac}        
\usepackage{microtype}       
\usepackage{xcolor}          
\usepackage{colortbl}
\usepackage{graphicx}
\usepackage{float}
\usepackage{tikz}
\usepackage{caption}
\usepackage{subcaption}
\usepackage{multirow}
\usepackage{algorithm}
\usepackage{algpseudocode}
\usepackage[normalem]{ulem}

\newcommand{\diag}{\operatorname{diag}}
\newcommand{\Real}{\operatorname{Re}}

\newcommand{\R}{\mathbb{R}}
\newcommand{\C}{\mathbb{C}}

\newcommand{\e}{\mathrm{e}}
\newcommand{\ii}{\mathrm{i}}

\newcommand{\rmF}{\mathrm{F}}

\newcommand{\bfb}{\boldsymbol{b}}
\newcommand{\bfc}{\boldsymbol{c}}

\newcommand{\bfh}{\boldsymbol{h}}
\newcommand{\bfp}{\boldsymbol{p}}
\newcommand{\bfs}{\boldsymbol{s}}
\newcommand{\bfu}{\boldsymbol{u}}

\newcommand{\bfy}{\boldsymbol{y}}

\newcommand{\bfA}{\boldsymbol{A}}
\newcommand{\bfq}{\boldsymbol{q}}
\newcommand{\bfo}{\boldsymbol{o}}
\newcommand{\bfB}{\boldsymbol{B}}

\newcommand{\bfW}{\boldsymbol{W}}
\newcommand{\bfg}{\boldsymbol{g}}

\newcommand{\bfH}{\boldsymbol{H}}

\newcommand{\bfzero}{\boldsymbol{0}}

\newcommand{\GopTheta}{\mathcal{G}_{\theta}}
\newcommand{\relL}{\mathrm{rel}\,L_{2}}

\makeatletter
\@addtoreset{equation}{section}

\makeatother

\title{ZNO: Stable Rational Neural Operators in the Z-Domain for Discrete-Time Dynamics}

\author{%
  Xianli Zhu\\
  School of Mathematical Sciences\\
  Fudan University\\
  Shanghai, China\\
  \texttt{25210180115@m.fudan.edu.cn}
  \And
  Jia Yin\thanks{Corresponding author: \texttt{jiayin@fudan.edu.cn}.}\\
  School of Mathematical Sciences\\
  Fudan University\\
  Shanghai, China\\
  \texttt{jiayin@fudan.edu.cn}
}

\begin{document}
\maketitle

\begin{abstract}
    We introduce the \emph{Z-Domain Neural Operator} (ZNO), a causal neural operator whose layers are stable low-rank multiple-input multiple-output (MIMO) rational filters parameterized directly in the $z$-plane. This operator is designed to address a critical limitation of existing operator learning methods, as most of these methods are primarily tailored for continuous-time problems, while a large class of system-identification problems is intrinsically discrete-time.
The $z$-domain form expresses stability as a unit-disk pole constraint and makes
learned discrete-time poles directly readable. The model combines low-rank channel
mixing, smooth stable pole reparameterization, causal recurrence, and an optional
short finite impulse response (FIR) branch in a single $z$-domain rational recurrent
layer. Across controlled discrete system-identification experiments, ZNO's advantage is
most evident when the target dynamics are stable rational systems with lightly damped
poles near the unit circle. Under matched parameter budgets, ZNO is not uniformly
dominant; however, with validation-selected configurations, the same architecture
can achieve the lowest mean error across the controlled tasks. A five-bin difficulty sweep over near-unit-circle / long-memory
dynamics further shows that ZNO has the lowest mean error across all memory
regimes, from short ($\approx\!10$ steps) to long 
($\approx\!100-200$ steps). On five public nonlinear system-identification
benchmarks, ZNO is competitive with neural operator and state-space baselines, achieving the lowest
mean error on benchmarks whose dynamics align with stable rational discrete-time
filters, while classical or state-space baselines remain preferable on some systems.
These results position ZNO as a strong model for stable rational discrete-time
dynamics, especially in near-unit-circle and long-memory regimes, but not as a
universal replacement for specialized system-identification methods.

\end{abstract}

\section{Introduction}
\label{sec:introduction}

Neural operators learn mappings between infinite-dimensional function spaces from
data~\cite{chen1995universal,kovachki2023nop,li2021fno,lu2021deeponet}. Two transform-based families
dominate the current landscape: the \emph{Fourier Neural Operator
(FNO)}~\cite{li2021fno}, which parameterizes kernel integrals by learnable Fourier
coefficients on the unit circle; and the \emph{Laplace Neural Operator
(LNO)}~\cite{cao2024lno}, which parameterizes each layer by continuous-time poles
and residues on the left-half $s$-plane. These parameterizations are most natural
when the target operator is defined over an underlying continuous-time or
continuous-space domain, as in ODE/PDE settings, and the discrete samples arise
from numerical discretization rather than from the native dynamics of the system.
In contrast, many systems of interest are either intrinsically discrete-time or
observed at a fixed sampling interval. Examples include autoregressive
moving-average (ARMA) models, digital infinite impulse response (IIR) / finite
impulse response (FIR) filter banks~\cite{oppenheim1999discrete}, discretized
control plants, nonlinear autoregressive with exogenous inputs (NARX)
identification problems~\cite{ljung1999system}, and discretized time-stepper
surrogates in scientific machine learning~\cite{kovachki2023nop}. For these
systems, the ground-truth solution operator should be a rational function of $z^{-1}$ on the unit
disk rather than a rational function of $s$ on the left-half plane, and stability
requires that the learned poles lie strictly inside the unit circle.

\begin{figure}[htbp]
    \centering
    \includegraphics[width=\linewidth]{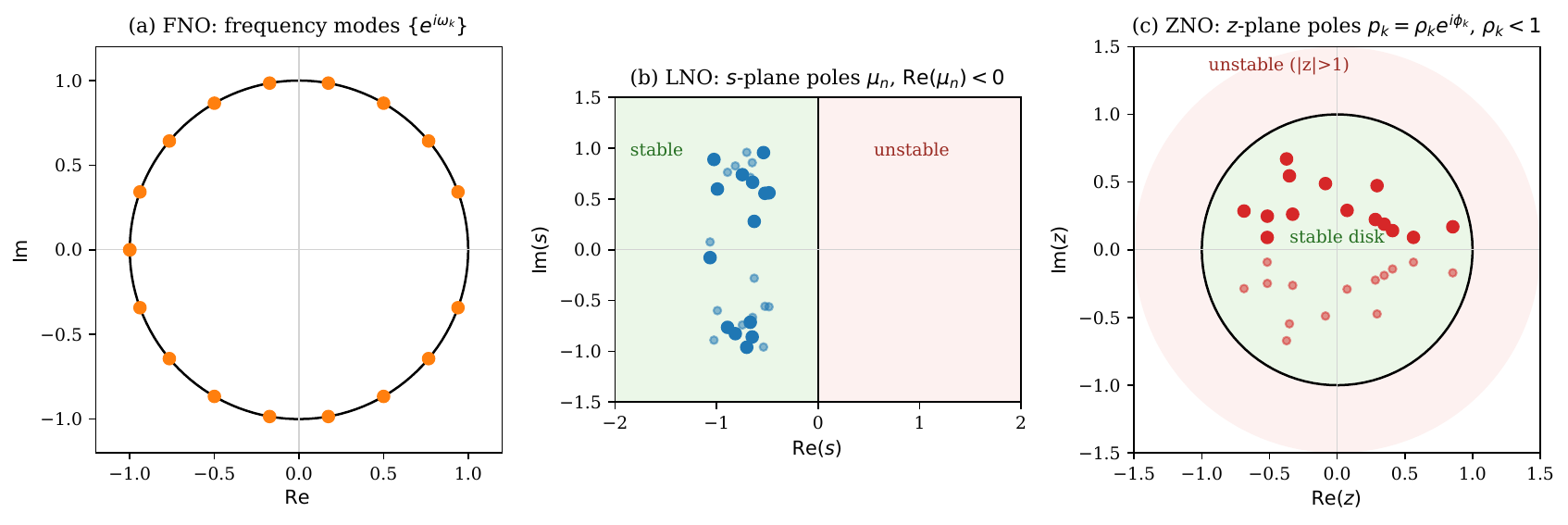}
    \caption{Spectral geometries of FNO, LNO and ZNO. (a) FNO parameterizes the
    transfer function at equally spaced points on the unit circle. (b) LNO
    parameterizes continuous-time poles in the left-half $s$-plane. (c) ZNO
    parameterizes discrete-time poles inside the unit disk of the $z$-plane; the
    unit circle is the natural stability boundary for sampled sequences.}
    \label{fig:stability_concept}
\end{figure}

\paragraph{Motivation.}
A continuous-time pole $\mu\in\C^{-}$ has stable impulse response $\e^{\mu t}$.
When sampled with step $\Delta t$, it becomes a discrete-time pole $p=\e^{\mu\Delta t}$
inside the unit disk. The LNO parameterization $\mu=-\alpha+\ii\beta$ with
$\alpha>0$ therefore gives $p=\e^{-\alpha\Delta t}\e^{\ii\beta\Delta t}$. However, when
the data are discrete-time from the start, there is no natural $\Delta t$ to invert. In this case, a continuous-time parameterization must instead adopt an arbitrary $\Delta t$, and then apply an exponential mapping that converts the sign constraint on $\Real(\mu)$ into
the magnitude bound $|p|<1$. Parameterizing poles directly in the $z$-plane therefore gives a natural
interface for discrete-time dynamics: stability is expressed by the unit disk,
and learned poles can be inspected directly in discrete-time coordinates.
Appendix~\ref{app:isomorphic} later separates this modelling choice from a
coordinate-only effect by comparing ZNO with an isomorphic $s$-plane
reparameterization of the same low-rank rational layer. That ablation shows that
the empirical gains should be attributed to the full stable rational recurrent
layer, rather than to the pole coordinate chart alone. Figure~\ref{fig:stability_concept} summarizes the
spectral geometries of FNO, LNO and ZNO.

\paragraph{Contributions.}
\begin{enumerate}
    \item We introduce the Z-Domain Neural Operator (ZNO), a stable rational neural
    operator whose basic building block is a low-rank multiple-input multiple-output
    (MIMO) rational filter in the $z$-domain. By low-rank MIMO, we mean that instead of allocating an independent
    rational filter to each input-output channel pair, the dynamic transfer from $w$
    hidden input channels to $w$ hidden output channels is routed through only
    $r\ll w$ latent scalar rational filters. The ZNO layer's learnable parameters
    are (i) complex poles constrained to the open unit disk by a smooth
    sigmoid, (ii) complex residues arranged in conjugate‑symmetric pairs to yield a real impulse response, and (iii) an optional short FIR branch. The recurrence can
    be implemented by a fused CUDA/Triton scan~\cite{tillet2019triton} with linear cost in sequence length.
    As this is an implementation acceleration instead of a modelling change, it does not
    affect the hypothesis class or training objective
    (Appendix~\ref{app:fused_kernel}).
    \item We construct a three-task discrete-time system-identification benchmark
    (resonant ARMA with poles near the unit circle, sixth-order IIR cascade, scalar
    NARX) and evaluate FNO~\cite{li2021fno}, the diagonal state-space model
    S4D~\cite{gu2022s4d}, and ZNO under a
    strictly matched parameter budget of $8.5-8.9$~k (5 seeds). ZNO achieves the lowest
    mean error on the flagship resonant task (Welch $t\!\approx\!-31$;
    $\relL\!=\!0.203$ vs. $0.386$ for S4D), narrowly leads on
    NARX ($0.309$ vs. $0.311$), and is second to S4D on IIR at the matched budget
    ($0.724_{\pm 0.138}$ vs.
    $0.704_{\pm 0.011}$). However, under the tuned-best protocol, ZNO has the lowest mean
    error on all three tasks.
    \item We identify in which ZNO achieves its advantage using a five-bin
    sweep across near-unit-circle / long-memory difficulty. The pole radius is varied
    over $[0.90,0.995]$, which controls the effective memory horizon from
    approximately $10$ to $200$ steps. ZNO has the lowest mean error in
    every bin. The performance gap between ZNO and S4D narrows but persists even in the
    most challenging long-memory regime (bin 4: ZNO $0.549_{\pm 0.015}$,
    S4D $0.735_{\pm 0.005}$, FNO $0.768_{\pm 0.001}$).
    \item We find out that an isomorphic $z$ vs.\ $s$ ablation within the same low-rank
    rational layer shows no practically meaningful difference at this sample size.
    Hence, the $z$-domain rational formulation is central to the model design, while the
    coordinate parameterization alone does not drive the observed gains. As a result, we
    attribute the accuracy gain \emph{jointly} to the rational low-rank layer, the
    smooth stable-pole reparameterization, and the causal recurrent structure. Both ZNO and S4D extrapolate
    stably to $4\times$ the training length, but ZNO retains the lowest absolute
    error on every task, while FNO's error grows by $1.9-3.6$ times.
    \item We calibrate external validity on five public nonlinear
    system-identification benchmarks loaded through the
    \texttt{nonlinear-benchmarks} package~\cite{beintema2025nonlinearbenchmarks}, using the released train/test splits, a
    sequential $80/20$ train/validation split on the training stream,
    train-stream normalization, and the benchmark-provided state-initialization
    windows. ZNO achieves the lowest mean error on Silverbox~\cite{wigren2013silverbox},
    Wiener--Hammerstein~\cite{schoukens2009wh} and the coupled electric drives
    (CED) benchmark~\cite{wigren2017ced}; S4D has the lowest mean error on cascaded
    tanks~\cite{schoukens2016tanks}; and a low-order classical
    system-identification model~\cite{lacerda2020sysidentpy} has the lowest error on the
    electro-mechanical positioning system (EMPS)~\cite{janot2019emps}. ZNO is therefore competitive
    but not universal on public system-identification data: its advantage
    concentrates in near-unit-circle, long-memory regimes and on stable rational
    discrete-time dynamics.
\end{enumerate}

\paragraph{Organization.}
Section~\ref{sec:preliminary} reviews the discrete-time operator learning and the
$z$-transform. Section~\ref{sec:architecture} defines the ZNO layer, its stable
pole-residue parameterization, and the fused recurrent kernel.
Section~\ref{sec:experiments} presents the main discrete benchmark, the
near-unit-circle difficulty sweep, long-horizon extrapolation, the suite of
public nonlinear system-identification benchmarks.
Section~\ref{sec:conclusion} concludes. The appendix contains training protocols,
the full isomorphic $z$-vs-$s$ ablation, the symmetric training-signal fairness
control, the 1D LNO ODE benchmark, the internal design-knob ablation, the IIR
stability follow-up, and the accuracy/wall-clock trade-off.

\section{Background}
\label{sec:preliminary}

\paragraph{Discrete-time operator learning.}
Let $T$ be the sequence length, $n\in\Omega=\{0,\dots,T-1\}$ the time index,
$d_u$ and $d_y$ the input and output channel dimensions, and $\theta$ the model
parameters. We use $\bfu\in\R^{T\times d_u}$ to denote a forcing and
$\bfy\in\R^{T\times d_y}$ to denote a response. Given $N$ i.i.d.\ training pairs
$\{(\bfu^{(j)},\bfy^{(j)})\}_{j=1}^{N}$, we learn a causal operator
$\GopTheta\colon\bfu\mapsto\bfy$, with $y_n=(\GopTheta\bfu)_n$ depending only on
$u_{0:n}$, by minimizing the batch-averaged unsquared relative Frobenius norm
\begin{equation}
    \label{eq:empirical_loss}
    \widehat{R}(\theta) \;=\;
    \frac{1}{N}\sum_{j=1}^{N}
    \frac{\|\GopTheta\bfu^{(j)} - \bfy^{(j)}\|_{\rmF}}{\|\bfy^{(j)}\|_{\rmF}},
\end{equation}
where $\|A\|_{\rmF}=(\sum_{n,c} A_{n,c}^{2})^{1/2}$ is the Frobenius norm over
the discretized time--channel array. Equivalently, this is the relative $L_2$
norm after flattening each sample, matching the reporting
convention in~\cite{cao2024lno,li2021fno}. The same metric is used at test time.

\paragraph{$z$-transform and pole-residue form.}
For a causal sequence $\{f_n\}_{n\geq 0}$, the $z$-transform is defined as 
$F(z)=\sum_{n\ge 0} f_n z^{-n}$~\cite{oppenheim1999discrete}. A
causal LTI transfer function admits the pole-residue decomposition
\begin{equation}
    \label{eq:pole_residue}
    G(z) \;=\; g_{\infty}\;+\;\sum_{k=1}^{K}\frac{c_k}{1-p_kz^{-1}}\;+\;\text{FIR}(z^{-1}),
\end{equation}
where $\text{FIR}(z^{-1})=\sum_{j=0}^{F}g_jz^{-j}$ is a finite impulse response branch
capturing direct feedthrough and short-memory effects, while the pole-residue terms
provide the long-memory IIR response. Bounded-input bounded-output (BIBO)
stability is equivalent to the condition $|p_k|<1$~\cite{oppenheim1999discrete}
for all $k$. For the $k$th summand in~\eqref{eq:pole_residue}, the impulse
response is $c_kp_k^n$ for $n\ge 0$. Hence, this mode decays exactly when
$|p_k|<1$.

\paragraph{Causal recurrence and state-space connection.}
Introducing a complex state $s_n^{(k)}$ for each pole $k=1,\ldots,K$,
\begin{equation}
    \label{eq:state_recurrence}
    s_n^{(k)} \;=\; p_k\,s_{n-1}^{(k)} \;+\; b_n,\qquad
    y_n \;=\; \Real\!\sum_{k=1}^{K} c_k\,s_n^{(k)} \;+\; (\text{FIR}\ast b)_n,
\end{equation}
realizes~\eqref{eq:pole_residue} pointwise in $n$, so the same parameters
$(p_k,c_k)$ define a valid operator at every sequence length $T'\ge 1$. The
recurrence~\eqref{eq:state_recurrence} is also the backbone of S4/S4D-family
state-space models~\cite{gu2022s4d,gu2021s4}. S4D parameterizes diagonal complex
state-space poles in continuous time and discretizes them to the $z$-plane through a
bilinear or zero-order-hold mapping. Therefore, we use S4D as our main
state-space baseline.
ZNO differs by parameterizing complex conjugate-pair poles directly on the $z$-plane,
factoring the MIMO rational layer into a low-rank form, and evaluating the rational
response through a causal recurrence. The fused kernel is an implementation
acceleration and does not change the recurrence being evaluated.

\section{The Z-Domain Neural Operator}
\label{sec:architecture}

\begin{figure}[htbp]
    \centering
    \includegraphics[width=\linewidth]{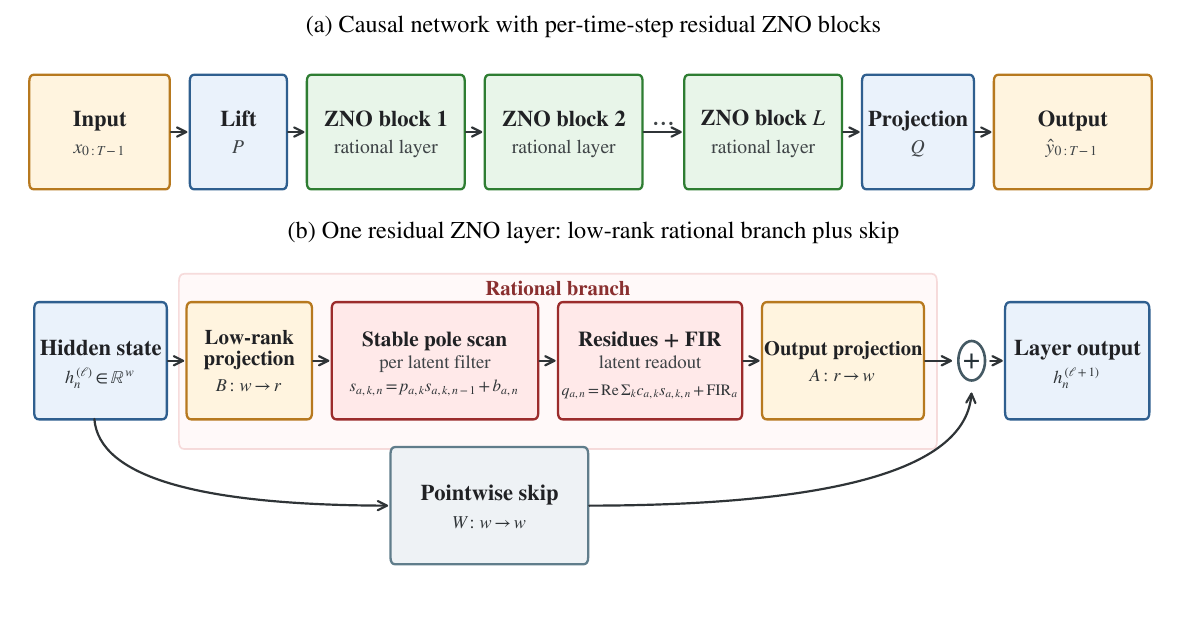}
    \caption{The ZNO architecture. (a) The full causal network: pointwise lift $P$, $L$
    residual ZNO layers with pointwise nonlinearities, pointwise projection $Q$. (b) One
    ZNO layer: the hidden sequence is projected to a low-rank stream; each latent rank
    channel is fed through its own bank of conjugate-pair stable poles, read out through
    learnable residues and an optional short FIR branch, projected back by $A:r\to w$,
    and added to the parallel pointwise skip $W h^{(\ell)}_n$.}
    \label{fig:arch}
\end{figure}

\paragraph{Network.}
Following the lift/transform/project template of FNO and LNO, a depth-$L$ ZNO network is
\begin{equation}
    \label{eq:zno_network}
    \bfy \;=\; Q \circ \sigma \circ \mathcal{T}^{(L)} \circ \cdots \circ
    \sigma\circ\mathcal{T}^{(1)} \circ P(\bfu),
\end{equation}
with pointwise lift $P$, projection $Q$, smooth pointwise nonlinearity $\sigma$
(GELU for discrete-native experiments), and residual layers
$\mathcal{T}^{(\ell)}=\mathcal{R}^{(\ell)}+\mathcal{W}^{(\ell)}$. Here
$\mathcal{R}^{(\ell)}$ is the low-rank MIMO rational branch defined below, while
$\mathcal{W}^{(\ell)}$ is a dense pointwise skip with matrix $\bfW^{(\ell)}$,
that is, $(\mathcal{W}^{(\ell)}\bfh)_n=\bfW^{(\ell)}\bfh_n$.
Throughout this paper, $w$ denotes the hidden width and $L$ denotes the network
depth; reported configurations use $w\in\{12,20\}$ and $L\in\{2,3,4\}$.
Figure~\ref{fig:arch} provides an illustrative explanation of this architecture.

\paragraph{Low-rank MIMO rational branch.}
The dynamic branch acts on $\bfh^{(\ell)}\in\R^{T\times w}$ via a low-rank
factorisation:
\begin{equation}
    \label{eq:lrlayer}
    \mathcal{R}^{(\ell)}(\bfh^{(\ell)})_n
    \;=\;
    \bfA^{(\ell)}\,\bfq^{(\ell)}_n,
    \quad
    \bfq^{(\ell)}_n
    =\Real\!\sum_{k=1}^{K} \bfc^{(\ell)}_{k}\odot\bfs^{(\ell,k)}_n
    + \mathbf{1}_{F>0}\sum_{j=0}^{F}\bfg^{(\ell)}_{j}\odot\bfb^{(\ell)}_{n-j},
\end{equation}
\begin{equation}
    \label{eq:state_recurrence_layer}
    \bfs^{(\ell,k)}_n \;=\; \bfp^{(\ell)}_{k}\odot\bfs^{(\ell,k)}_{n-1} \;+\; \bfb^{(\ell)}_n,
    \qquad
    \bfb^{(\ell)}_n \;=\; \bfB^{(\ell)}\,\bfh^{(\ell)}_n,
    \qquad
    \bfs^{(\ell,k)}_{-1}=\bfzero,
\end{equation}
where $\bfB^{(\ell)}\in\R^{r\times w}$, $\bfA^{(\ell)}\in\R^{w\times r}$,
$\bfW^{(\ell)}\in\R^{w\times w}$ is the matrix of the skip operator
$\mathcal{W}^{(\ell)}$,
$\bfp^{(\ell)}_k,\bfc^{(\ell)}_k\in\C^{r}$, $\bfg^{(\ell)}_j\in\R^{r}$, and
$\odot$ is the entrywise product. Each latent rank channel therefore has its own
scalar pole-residue filter. The pole count $K$ denotes total poles per latent filter.
Since the implementation uses $\lfloor K/2\rfloor$ conjugate pairs and, if $K$ is odd, one
additional real pole, the sum is real by construction. The FIR sum is omitted when
$F=0$ and otherwise covers lags $0,\ldots,F$. Intuitively, ZNO routes the
$w$-channel hidden sequence through $r\ll w$ latent scalar rational filters and then
mixes the filtered latent streams back to width $w$. Equivalently, ignoring the
pointwise nonlinearity, the dynamic transfer factors as
\[
    \bfH^{(\ell)}(z)
    =
    \bfA^{(\ell)}
    \diag\!\left(G^{(\ell)}_1(z),\ldots,G^{(\ell)}_r(z)\right)
    \bfB^{(\ell)}.
\]
Thus the MIMO dynamics are diagonal in the latent filter space and dense only through
the input/output mixing matrices $\bfB^{(\ell)}$ and $\bfA^{(\ell)}$, rather than
assigning an independent recurrent filter to every input-output channel pair.
Across the reported main, public, ablation and 1D checks,
ZNO sweeps use $r\in\{4,8,12,16\}$, $K\in\{8,16,20,24,32,40,48,64,80\}$ and
$F\in\{0,2,4,8\}$. The layer's parameter count and a comparison with FNO/S4D layers are given in
Appendix~\ref{app:parameter_count}.

\paragraph{Stable pole parameterization.}
Each pole entry is stored as $(\tilde\rho^{(\ell)}_{a,k},\phi^{(\ell)}_{a,k})\in\R^{2}$ for
latent channel $a$ and pole index $k$, and mapped by
\begin{equation}
    \label{eq:stable_reparam}
    \rho^{(\ell)}_{a,k} =
    \rho_{\max}\operatorname{sigmoid}(\tilde\rho^{(\ell)}_{a,k}),
    \qquad
    p^{(\ell)}_{a,k} = \rho^{(\ell)}_{a,k}\,\e^{\ii\phi^{(\ell)}_{a,k}},
\end{equation}
where $\operatorname{sigmoid}(x)=(1+\exp(-x))^{-1}$ and
$\rho_{\max}=0.999$ is a fixed hard stability ceiling. This smooth map sends
the unconstrained parameters onto $(0,\rho_{\max})$, enforces the unit-disk
constraint, and eliminates the need for a hard projection step. We introduce a smooth
pole-safety regulariser
\begin{equation}
    \label{eq:safety}
    \mathcal{R}_{\mathrm{pole}}(\theta)
    \;=\;\frac{1}{LrK}\sum_{\ell,a,k}\bigl(\max(|p^{(\ell)}_{a,k}|-\rho_{\mathrm{safe}},0)\bigr)^{2}
\end{equation}
with $\rho_{\mathrm{safe}}=0.95$. For ZNO, the training objective is
\begin{equation}
    \label{eq:zno_objective}
    \mathcal{L}_{\mathrm{ZNO}}(\theta)
    =
    \widehat{R}(\theta)
    +\lambda_{\mathrm{pole}}\mathcal{R}_{\mathrm{pole}}(\theta)
    +\lambda_{\mathrm{suf}}\widehat{R}_{\mathrm{suf}}(\theta),
\end{equation}
where $\widehat{R}$ is the relative Frobenius loss in~\eqref{eq:empirical_loss} and
\[
    \widehat{R}_{\mathrm{suf}}(\theta)
    =
    \frac{1}{N}\sum_{j=1}^{N}
    \frac{\|(\GopTheta\bfu^{(j)}-\bfy^{(j)})_{T_{\mathrm{ctx}}:T-1}\|_{\rmF}}
         {\|\bfy^{(j)}_{T_{\mathrm{ctx}}:T-1}\|_{\rmF}},
    \qquad
    T_{\mathrm{ctx}}=\lfloor T/4\rfloor .
\]
We use $\lambda_{\mathrm{pole}}=10^{-3}$ and $\lambda_{\mathrm{suf}}=10^{-2}$.
The suffix term is computed from the same full forward pass and is not an
autoregressive evaluation. These two auxiliary terms are applied only inside the
ZNO branch (and its $s$-plane-isomorphic sibling), while FNO and S4D are trained with the
plain relative $L_2$ loss alone. A symmetric training-signal fairness control
(Section~\ref{sec:experiments_discrete}, Appendix~\ref{app:suffix_fairness}) shows
that this asymmetry does not account for the advantage of ZNO.

\paragraph{Fused recurrent kernel.}
We implement the recurrence~\eqref{eq:state_recurrence_layer} as a fused
CUDA/Triton~\cite{tillet2019triton} scan. The fused path computes the same
recurrence as the PyTorch reference and changes neither the hypothesis class, the
training objective, nor the predictions beyond numerical roundoff. Scan
pseudocode, backward-memory trade-offs, timing, and the relation to an LNO
reparameterization are provided in Appendix~\ref{app:fused_kernel}.

\section{Experiments}
\label{sec:experiments}

In this section, we evaluate ZNO across synthetic and public discrete-time
system-identification tasks and long-horizon extrapolation tests.
The training protocol, the symmetric training-signal
fairness control, the 1D LNO ODE benchmark, the internal ablation, the
accuracy-vs-wall-clock trade-off, pole-map diagnostics and qualitative prediction traces are
deferred to
Appendices~\ref{app:protocol}-\ref{app:traces}.

\paragraph{Setup.}
All experiments were conducted on a single NVIDIA A30 (24\,GB) with PyTorch~\cite{paszke2019pytorch} and CUDA.
Every model is trained with Adam, weight decay $10^{-4}$, a StepLR schedule with
decay factor $\gamma=0.5$, batch size $32$, and $600$ epochs on the discrete
benchmark; we report the best-validation checkpoints. For the discrete-benchmark data, we used
$N_{\mathrm{train}}/N_{\mathrm{val}}/N_{\mathrm{test}}=1024/256/256$ i.i.d.\
trajectories of length $T=2048$. All discrete-benchmark tables below use five
seeds per cell; the 1D LNO ODE benchmark in
Appendix~\ref{app:1d_benchmark} uses three seeds. Further details, per-task
configurations and asset traceability are provided in Appendix~\ref{app:protocol}.

\subsection{Discrete-time system-identification benchmark}
\label{sec:experiments_discrete}

This experiment compares FNO, S4D and ZNO on three controlled discrete-time
system-identification tasks under matched-budget and tuned-best protocols. The
benchmark contains two parameter-conditioned sequence-to-sequence tasks and one
unconditioned input-output task.
\begin{itemize}
    \item \textbf{Resonant ARMA (near unit circle).} Second-order autoregressive
    plant with a conjugate pole pair $p=\rho\,\e^{\ii\phi}$,
    $\rho\sim\mathcal{U}(0.9,0.995)$, $\phi\sim\mathcal{U}(0.05\pi,0.45\pi)$, and
    a 4-tap MA numerator, driven by colored noise. Input: $7$ channels (forcing
    + six system parameters).
    \item \textbf{Sixth-order IIR cascade.} Three biquad sections with independent
    resonant poles in $(0.88,0.995)$. Input: $16$ channels (forcing + five
    scalars per biquad).
    \item \textbf{Nonlinear NARX (softplus and tanh).} A scalar unconditioned
    NARX recurrence
    $y_n = a_1\,y_{n-1} + a_2\,y_{n-2}
          + \mathrm{gain}\,\bigl(\log(1+\e^{x_n+0.5\,x_{n-1}}) - \log 2\bigr)
          + 0.08\,\tanh(y_{n-1}\,x_n)$,
    with $a_1\!\sim\!\mathcal{U}(0.2,0.6)$, $a_2\!\sim\!\mathcal{U}(-0.3,0.1)$,
    and $\mathrm{gain}\!\sim\!\mathcal{U}(0.15,0.45)$ sampled per trajectory but
    \emph{not} exposed to the model. Input: $1$ channel (the forcing); the
    operator performs system identification from the input-output trace alone.
\end{itemize}
The two parameter-conditioned tasks are closer to parameter-aware system
identification~\cite{ljung1999system} than to the classical operator-learning
setting, where only the forcing input varies. In contrast, NARX is a pure input-output
identification problem.

We compare ZNO with FNO~\cite{li2021fno} which is a frequency-domain neural operator
and with S4D~\cite{gu2022s4d} which is a state-space model. A pilot run of the
reference LNO layer on an earlier four-task discrete pilot diverged on every
case, so LNO is excluded from the discrete benchmark
(Appendix~\ref{app:lno_pilot}). Under strictly matched configurations, FNO,
S4D and ZNO each operate within a parameter range of $8457-8917$ parameters on every task
(Table~\ref{tab:matched_configs} in Appendix~\ref{app:protocol}).

\begin{figure}[htbp]
    \centering
    \includegraphics[width=\linewidth]{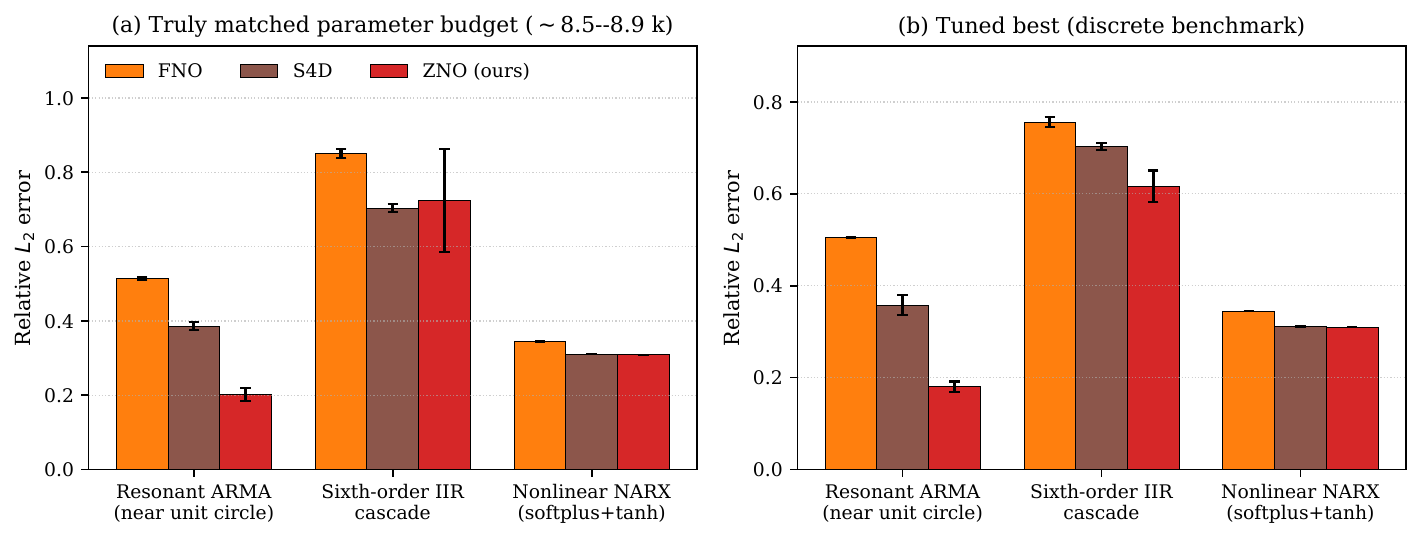}
    \caption{Discrete benchmark results: relative $L_2$ test error
    (mean\,$\pm$\,standard deviation, 5 seeds). (a) Matched-budget protocol:
    all models use $8.5$--$8.9$k parameters on every task. (b) Tuned-best
    protocol. ZNO has the lowest mean error on the near-unit-circle resonant
    task and narrowly on NARX, but is not uniformly dominant under the matched
    budget.}
    \label{fig:benchmark_discrete}
\end{figure}

\begin{table}[htbp]
    \centering
    \small
    \begin{tabular}{@{}llrrr@{}}
        \toprule
        Task & Protocol & FNO & S4D & ZNO (ours)\\
        \midrule
        Resonant ARMA      & matched & $0.515_{\pm 0.004}$ & $0.386_{\pm 0.011}$  & $\mathbf{0.203_{\pm 0.017}}$\\
        Resonant ARMA      & tuned   & $0.505_{\pm 0.001}$ & $0.358_{\pm 0.022}$  & $\mathbf{0.181_{\pm 0.011}}$\\
        IIR cascade (6th)  & matched & $0.850_{\pm 0.013}$ & $\mathbf{0.704_{\pm 0.011}}$ & $0.724_{\pm 0.138}$\\
        IIR cascade (6th)  & tuned   & $0.756_{\pm 0.011}$ & $0.703_{\pm 0.007}$  & $\mathbf{0.616_{\pm 0.035}}$\\
        Nonlinear NARX     & matched & $0.345_{\pm 0.001}$ & $0.311_{\pm 0.0002}$ & $\mathbf{0.309_{\pm 0.0001}}$\\
        Nonlinear NARX     & tuned   & $0.345_{\pm 0.001}$ & $0.311_{\pm 0.0004}$ & $\mathbf{0.309_{\pm 0.00004}}$\\
        \bottomrule
    \end{tabular}
    \caption{Discrete benchmark results: relative $L_2$ test error
    (mean\,$\pm$\,standard deviation, 5 seeds). The rows labeled ``Matched'' use
    the symmetric parameter budget across all models, while the rows labeled
    ``Tuned'' use each model's validation-selected discrete configuration
    (Appendix~\ref{app:protocol}).
    The lowest mean error in each row is shown in bold. Under both protocols, ZNO achieves the lowest mean error on the resonant ARMA task by a wide margin and on the NARX task by a small margin. At the matched budget on IIR, S4D yields the lowest-error baseline; however, 
    under the tuned-best configurations, ZNO attains the lowest mean error on all three
    tasks.}
    \label{tab:discrete_main}
\end{table}

\paragraph{Results.}
Figure~\ref{fig:benchmark_discrete} and Table~\ref{tab:discrete_main} summarize the
key findings. The ZNO advantage is most pronounced when the target dynamics are
governed by lightly damped poles near the unit circle, smaller on the
noise-limited NARX task, and less consistent on the sixth-order IIR cascade under
the shared optimizer. The clearest advantage appears on the resonant ARMA task, where
the target dynamics are dominated by near-unit-circle poles: the five ZNO seeds yield errors in
$\{0.189,0.186,0.191,0.221,0.226\}$, while the five S4D seeds average $0.386$ with
standard deviation $0.011$ (Welch $t\!\approx\!-31$). On NARX, all three models approach a
noise-limited floor near $\relL\!\approx\!0.31$. The ordering
ZNO~<~S4D~<~FNO is clean (Welch $t\!\approx\!-22$), but the absolute margin is
small. On the sixth-order IIR cascade, S4D reaches $0.704_{\pm 0.011}$, slightly
below the ZNO mean $0.724_{\pm 0.138}$. The large ZNO standard deviation is due to a single divergent seed: four of five seeds converge close to $\{0.650,0.663,0.653,0.656\}$,
while one diverges to $\relL=1.000$. A follow-up stability experiment with a stronger gradient
clip (Appendix~\ref{app:iir_stability}) brings ZNO to $0.706_{\pm 0.023}$ and its
isomorphic $s$-plane variant to $0.693_{\pm 0.014}$, indicating that the divergence is an
optimization artifact of the low-rank rational layer rather than a
$z$-plane-specific pathology. Under the tuned-best protocol ZNO
($0.616_{\pm 0.035}$) attains the lowest mean error on this task. Finally, a symmetric
training-signal fairness control using suffix-loss-matched baselines and disabling ZNO's auxiliary terms leaves the ordering unchanged across all tasks
(Appendix~\ref{app:suffix_fairness}).

\subsection{Near-unit-circle difficulty scaling}
\label{sec:experiments_difficulty}

This experiment isolates how pole radius, and therefore memory horizon, affects
the comparison between FNO, S4D and ZNO. The main resonant-ARMA task draws poles
from $\rho\in(0.9,0.995)$, thereby mixing short- and long-memory dynamics. To isolate the regime
that drives the ZNO advantage, we construct five narrower pole-radius bins, each controlling a narrow range of effective memory horizons. Within a given bin, the dominant impulse-response time constant is well approximated by
$\tau\approx-1/\log\rho$. The bins span $\rho\!\in\!\{[0.90,0.93], [0.93,0.96],
[0.96,0.98], [0.98,0.99], [0.99,0.995]\}$, corresponding to memory horizons of approximately $\tau\in\{ 10-14, 14-25, 25-50, 50-100,
100-200\}$ steps at sequence length $T=2048$. For each bin, we perform a fresh five-seed run of FNO, S4D and ZNO using the tuned-best configurations reported in
Table~\ref{tab:tuned_configs}.

\begin{figure}[htbp]
    \centering
    \includegraphics[width=\linewidth]{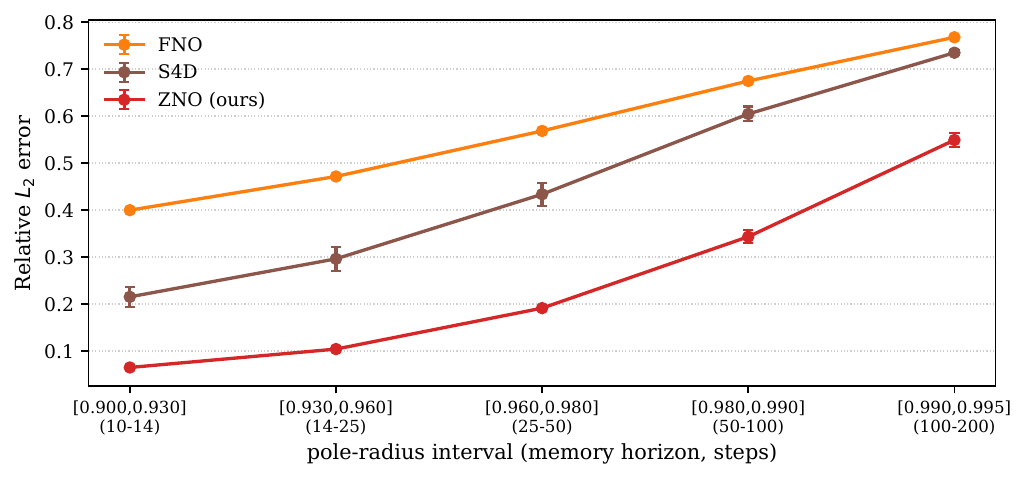}
    \caption{Near-unit-circle difficulty sweep. The pole-radius intervals
    control the effective memory horizon, given by $\tau\!\approx\!-1/\log\rho$. This horizon grows
    from roughly $10$ steps for $\rho\!\in\![0.90,0.93]$ to approximately $100-200$
    steps for $\rho\!\in\![0.99,0.995]$. Error bars indicate the standard deviation over five seeds. ZNO achieves the
    lowest mean error of the three models in every bin. The performance gap relative to S4D narrows
    in the most challenging long-memory regime but does not close entirely.}
    \label{fig:difficulty}
\end{figure}

\paragraph{Results.}
Figure~\ref{fig:difficulty} shows three main findings. First, as expected, every model becomes more challenging as the
memory horizon increases. Second, ZNO achieves the lowest mean error in
every bin, ranging from the easiest short-memory regime (bin 0: ZNO
$0.065_{\pm 0.004}$ vs S4D $0.215_{\pm 0.021}$, FNO $0.400_{\pm 0.002}$) to
the hardest long-memory regime (bin 4: ZNO $0.549_{\pm 0.015}$ vs S4D
$0.735_{\pm 0.005}$, FNO $0.768_{\pm 0.001}$). Third, the ZNO-to-S4D ratio
grows from $0.30$ on the easiest bin to $0.75$ on the hardest bin. In other words, the absolute advantage of ZNO
diminishes as the memory horizon becomes very long,
yet ZNO still maintains the lowest mean error throughout. The advantage is present across the
entire near-unit-circle sweep, but it is most pronounced in the short- to mid-memory bins
and narrows in the most challenging long-memory regime. These results support the view that ZNO is
particularly effective for stable rational discrete-time dynamics with near-unit-circle
poles, while very long memory remains challenging for all models.

\subsection{Long-horizon extrapolation}
\label{sec:experiments_extrap}

This experiment tests whether models trained at length $T=2048$ remain stable
when evaluated on longer sequences. We evaluate the tuned-best checkpoints of
Section~\ref{sec:experiments_discrete} on test sequences of length
$T'\in\{4096,8192\}$ (i.e., $2\times$ and $4\times$ the training length),
using five seeds per cell. Both ZNO and S4D are
essentially length-invariant because their layers are causal recurrences with
parameter counts that do not depend on sequence length. Specifically, the ratio of test error at $8192$ to that at $2048$ is 
$\{0.90,1.06,1.05\}$ for ZNO on \{resonant ARMA, IIR, NARX\}, and
$\{0.96,1.04,1.07\}$ for S4D. ZNO maintains the lowest absolute error on every task at
every evaluated length. In contrast, FNO's error grows substantially, by $3.6\times$, $2.7\times$, $1.9\times$ on resonant
ARMA, IIR and NARX at $T=8192$. This is consistent with its frequency grid being tied
to the training length. Full numerical results are provided in Table~\ref{tab:extrapolation}
(Appendix~\ref{app:extrap_full}). A parallel long-horizon evaluation across the
five difficulty bins (Appendix~\ref{app:extrap_full}) shows the same pattern,
regime by regime.

\subsection{Public nonlinear system-identification benchmarks}
\label{sec:experiments_public}

This experiment evaluates external validity on public nonlinear
system-identification benchmarks whose structure is not designed around ZNO. The
three tasks in Section~\ref{sec:experiments_discrete} are synthetic systems
chosen to probe rational-filter inductive bias; here we evaluate five public nonlinear system-identification
benchmarks~\cite{janot2019emps,schoukens2009wh,schoukens2016tanks,wigren2013silverbox,wigren2017ced}
loaded through the \texttt{nonlinear-benchmarks} package~\cite{beintema2025nonlinearbenchmarks}, including Silverbox (a Duffing-like electronic circuit), Wiener--Hammerstein (a static nonlinearity
sandwiched between two linear dynamic blocks), cascaded tanks (a fluid-flow
benchmark with overflow nonlinearity), coupled electric drives (CED), and the
electro-mechanical positioning system (EMPS). All datasets are evaluated with the official
train/test split, an $80/20$ sequential train/validation split on the training
stream, train-stream normalization, and the benchmark-provided
state-initialization prefix excluded from the reported error
(Appendix~\ref{app:public_protocol}). For each neural architecture, we select
the configuration by validation error on the public benchmark's training split,
and report five-seed confirmation results for the chosen configuration. The
classical column reports the best validation-selected low-order NARMAX model
from a pool of four AOLS/FROLS polynomial and Fourier
baselines~\cite{lacerda2020sysidentpy}.

\begin{table}[htbp]
    \centering
    \footnotesize
    \setlength{\tabcolsep}{5pt}
    \begin{tabular}{@{}lrrrr@{}}
        \toprule
        Benchmark & best classical & FNO & S4D & ZNO (ours)\\
        \midrule
        Silverbox            & $0.041$          & $0.984_{\pm 0.051}$ & $0.060_{\pm 0.011}$ & $\mathbf{0.021_{\pm 0.001}}$\\
        Wiener--Hammerstein  & $0.187$          & $0.033_{\pm 0.002}$ & $0.030_{\pm 0.007}$ & $\mathbf{0.003_{\pm 0.0001}}$\\
        CED                  & $0.221$          & $0.204_{\pm 0.022}$ & $0.177_{\pm 0.035}$ & $\mathbf{0.150_{\pm 0.022}}$\\
        Cascaded tanks       & $0.480$          & $0.382_{\pm 0.056}$ & $\mathbf{0.324_{\pm 0.014}}$ & $0.370_{\pm 0.037}$\\
        EMPS                 & $\mathbf{0.080}$ & $0.321_{\pm 0.040}$ & $0.338_{\pm 0.007}$ & $1.041_{\pm 0.147}$\\
        \bottomrule
    \end{tabular}
    \caption{Public nonlinear system-identification benchmarks. Neural network results
    are reported as mean\,$\pm$\,standard deviation over five seeds, following validation-based configuration
    selection. The classical column reports the best low-order NARMAX model
    selected on the same validation split. The lowest error on
    each benchmark is shown in bold.}
    \label{tab:public_main}
\end{table}

\paragraph{Results.}
Table~\ref{tab:public_main} places ZNO in a realistic comparative context. ZNO achieves the
lowest mean error on Silverbox, Wiener--Hammerstein and CED. S4D performs best on cascaded tanks, while a low-order classical system-identification
model has the lowest error on EMPS, substantially below all neural models on that
benchmark. On EMPS,
ZNO also exhibits a large validation/test gap: the best-validation error is approximately
$0.22$ compared to a test error of $1.04$. We keep the validation-selected configuration
fixed and report this gap directly, preserving the public test split as an external
validity check. Appendix~\ref{app:public_protocol} provides the validation/test gaps
for all five benchmarks (Table~\ref{tab:public_val_test_gap}). These results place a
useful boundary around our claims: ZNO is most effective when the observed dynamics are
well approximated by stable rational discrete-time filters. In contrast, low-order
physically identifiable systems, such as EMPS, can still favor classical
system-identification models; and benchmarks whose dynamics align better with a diagonal
continuous state-space bias, such as Cascaded tanks, can favor S4D.
Selected configurations and parameter counts are reported in
Appendix~\ref{app:public_protocol}.
Pole-map diagnostics and qualitative traces are provided in
Appendix~\ref{app:traces}.

\section{Conclusion}
\label{sec:conclusion}

ZNO is a stable rational neural operator whose elementary building block is a
low-rank MIMO rational filter with smooth $z$-plane pole stability. A fused GPU
scan implements the recurrence without changing the model class. Under a strictly
symmetric matched budget, ZNO achieves the lowest mean error on the flagship resonant
task (Welch $t\!\approx\!-31$; $\relL=0.203$ vs $0.386$ for S4D) and narrowly
outperforms S4D on NARX. On the sixth-order IIR cascade, S4D attains
$0.704$, slightly below the main-protocol ZNO mean of $0.724$ (the latter is inflated by one divergent seed, which a stability follow‑up resolves symmetrically for both pole parameterizations). Under the tuned-best protocol, ZNO has the lowest mean error
on all three tasks. Both ZNO and S4D extrapolate stably to $4\times$ the
training length. Across the evaluated lengths, ZNO retains the lowest absolute
error on every task, whereas FNO degrades more noticeably, with error growth of
$1.9$--$3.6\times$ at $T'=8192$. The $z$-domain rational formulation is central to this model
design; however, the isomorphic $z$ vs.\ $s$ ablation shows no practically meaningful
difference at this sample size when the low-rank rational recurrent layer is fixed.
We therefore attribute the accuracy gain to the full rational recurrent layer rather
than to the coordinate parameterization alone.

\textbf{Where the advantage lives.} A five-bin difficulty sweep over near-unit-circle /
long-memory pinpoints the regime in which ZNO outperforms both FNO and
S4D. ZNO achieves the lowest mean error in every bin, from short memory
($\approx\!10$ steps,
$\relL=0.065$) to the hardest long-memory bin ($\approx\!100-200$ steps,
$\relL=0.549$). On five public nonlinear system-identification benchmarks,
evaluated with the official splits, train-stream normalization, and the
benchmark-provided state-initialization windows, ZNO attains the lowest mean error
on Silverbox,
Wiener--Hammerstein and CED (the three benchmarks that are plausibly dominated
by stable rational discrete-time dynamics). S4D performs best on
Cascaded tanks, while a low-order classical system-identification model yields the
lowest error on EMPS.
Consequently, ZNO should be viewed as a stable rational operator for discrete-time,
near-unit-circle, and long-memory regimes, rather than as a universal replacement for
classical system-identification pipelines. Its advantage is most pronounced when the target
dynamics are well approximated by stable rational discrete-time filters.

\textbf{Limitations.} (i) Under the shared main-protocol optimizer and matched parameter budget, ZNO does not achieve the best performance on the sixth-order IIR cascade. A stability follow-up
(Appendix~\ref{app:iir_stability}) eliminates the divergent ZNO seed with a
modified optimizer recipe that we do not promote into the main table. (ii) On the EMPS benchmark, ZNO
exhibits a large validation/test gap (Table~\ref{tab:public_val_test_gap}), and a
structured classical model outperforms all neural baselines. This marks an
explicit limitation of the current ZNO model class on this benchmark. We keep the
validation-selected configuration fixed and report the gap directly. (iii) Wall-clock comparisons reflect
practical implementation costs (a fused Triton scan for ZNO vs plain PyTorch for both
baselines), not FLOP-level counts. Future work should pair the unit-disk pole
constraint with formal stability or robustness guarantees, and extend the layer
beyond one-dimensional temporal systems to spatio-temporal operator learning.

\bibliographystyle{plain}
\bibliography{bib/references}

\appendix
\raggedbottom
\section{Parameter count of the ZNO layer}
\label{app:parameter_count}

For the even pole counts used by the reported configurations, layer $\ell$ has $rw$
(in-projection) $+$ $rw$ (out-projection) $+$ $w^2+w$ (skip) $+$ $Kr$ (complex-residue
real and imaginary parameters) $+$ $Kr$ (pole radii and phases) $+$
$\mathbf{1}_{F>0}r(F+1)$ (FIR) parameters. Across all our experiments this keeps a ZNO
layer strictly lighter than a comparable spectral convolution layer in FNO (which scales
as $\mathrm{modes}\cdot w^2$) and competitive with a diagonal S4D layer.

\section{Fused recurrent kernel and relation to LNO}
\label{app:fused_kernel}

\paragraph{Scan pseudocode.}
Algorithm~\ref{alg:scan} gives the forward scan for one ZNO layer (complex conjugate
pairs, no FIR). The recurrence is sequential in $n$ but parallel over batch size
$B_{\mathrm{batch}}$, rank $r$, and pole count $K$. A PyTorch-level loop of length
$T$ therefore incurs many kernel launches and dominates per-epoch wall time for
$T\ge 512$. The fused implementation admits two exact backward modes:
\texttt{save\_history} stores every intermediate state at
$O(B_{\mathrm{batch}}TrK)$ memory and performs one additional sequential pass;
\texttt{recompute} stores only the final state, reruns the forward scan, and
performs two sequential passes at $O(B_{\mathrm{batch}}rK)$ memory. We verified the
fused path against a PyTorch reference implementation, obtaining a maximum
absolute deviation of order $10^{-8}$ on both forward and gradient. On a single
NVIDIA A30 (24\,GB) the fused scan is $6$--$8\times$ faster than the reference
Python scan at $T=2048$, $r=12$, $K=64$, $B_{\mathrm{batch}}=32$. This kernel is an
implementation acceleration only: it computes the same recurrence as the
reference path and does not change the hypothesis class, training objective, or
predictions beyond numerical roundoff. Because the FNO and S4D baselines run
through their reference PyTorch implementations without comparable custom-kernel
optimization, the wall-clock comparisons should be interpreted as practical
implementation costs rather than FLOP-level fairness claims.

\begin{algorithm}[H]
\caption{Forward scan for one ZNO layer (complex conjugate pairs, no FIR).}
\label{alg:scan}
\begin{algorithmic}[1]
\Require hidden sequence $\bfh\in\R^{B_{\mathrm{batch}}\times T\times w}$, poles
    $\{\bfp_k\}_{k=1}^{K/2}\subset\C^r$, residues $\{\bfc_k\}_{k=1}^{K/2}\subset\C^{r}$,
    in-proj $\bfB$, out-proj $\bfA$, skip $\bfW$
\State $\bfb \gets \bfh\,\bfB^\top$ \Comment{$B_{\mathrm{batch}}\times T\times r$, real}
\State $\bfs_0 \gets \mathbf{0} \in \C^{B_{\mathrm{batch}}\times r\times (K/2)}$
\For{$n=0,\dots,T-1$}
    \For{each pole pair $k=1,\dots,K/2$}
        \State $\bfs^{(k)}_{n} \gets \bfp_k \odot \bfs^{(k)}_{n-1} + \bfb_{n}$
    \EndFor
    \State $\bfq_n \gets 2\,\Real\!\Bigl(\sum_k \bfc_k \odot \bfs^{(k)}_{n}\Bigr)$
    \State $\bfo_n \gets \bfA\,\bfq_n + \bfW\,\bfh_{n}$
\EndFor
\State \Return $\bfo\in\R^{B_{\mathrm{batch}}\times T\times w}$
\end{algorithmic}
\end{algorithm}

\paragraph{Timing protocol.}
All wall-clock numbers are collected under the following protocol: (i) the Triton
kernel backend; (ii) the fused save-history backward mode; (iii) the Triton
extension is pre-compiled before the timed region starts; (iv) the first CUDA
forward/backward step is executed as warm-up and excluded from the timer; (v) the
reported training time is the wall clock of the epoch loop only (excludes data
loading, model construction, checkpointing and test evaluation).

\paragraph{Relation to an LNO reparameterization.}
A natural question is whether ZNO coincides with LNO under the variable change
$\mu\leftrightarrow(\log p)/\Delta t$. Three differences make the constructions
distinct in practice.
\begin{enumerate}
    \item \textbf{Direct $z$-plane parameterization.} For a continuous-time pole
    $\mu$, the discrete pole is $p=\exp(\mu\Delta t)$, so an $s$-plane
    implementation must choose a sampling interval $\Delta t$ before evaluating
    the discrete recurrence. For native discrete-time data this interval is an
    interface choice rather than a physical modeling variable. ZNO parameterizes
    $p$ directly in the unit disk via~\eqref{eq:stable_reparam}, enforcing
    stability as $|p|<1$ and making the learned poles directly readable in the
    discrete-time domain.
    \item \textbf{Low-rank MIMO rational layer.} The reference LNO layer applies a
    dense per-channel multiplication in the Laplace domain. The ZNO dynamic
    branch~\eqref{eq:lrlayer}--\eqref{eq:state_recurrence_layer} is a low-rank
    factorization of a MIMO rational transfer matrix with $r\ll w$ state channels,
    and the full residual layer adds a dense pointwise skip. Under matched
    parameter budget this shifts capacity from spectral resolution to
    architectural compactness. This effect is quantified in
    Appendix~\ref{app:ablation}.
    \item \textbf{Fused causal recurrent implementation.} The reference LNO evaluates
    pole-residue terms using FFTs and dense frequency-by-pole contractions over
    the time grid and retained modes. In contrast, ZNO evaluates fixed-rank,
    fixed-pole filters by a causal recurrent scan whose cost is linear in $T$
    for fixed $r$ and $K$.
\end{enumerate}

\section{Training protocol and per-task configurations}
\label{app:protocol}

\paragraph{Hyper-parameters.}
All models are trained with Adam, weight decay $10^{-4}$, a StepLR schedule
($\gamma=0.5$, step size $s\in\{100,150\}$ as tabulated in
Tables~\ref{tab:matched_configs} and~\ref{tab:tuned_configs}), batch size $32$.
Discrete runs use $600$ epochs; the 1D LNO benchmark uses the per-case epoch
budgets of the original LNO release. The primary data loss is the batch-averaged
unsquared relative Frobenius loss, equivalently flattened relative $L_2$,
defined in~\eqref{eq:empirical_loss}. ZNO
additionally uses the pole-safety regularizer~\eqref{eq:safety} with
$(\rho_{\mathrm{safe}}=0.95,\lambda_{\mathrm{pole}}=10^{-3})$ and a suffix
relative-$L_2$ term with $(\lambda_{\mathrm{suf}}=10^{-2},T_{\mathrm{ctx}}=\lfloor T/4\rfloor)$.
A symmetric control that gives FNO and S4D the same suffix loss and turns the ZNO
auxiliary terms off is reported in Appendix~\ref{app:suffix_fairness}.

\paragraph{Data splits and seed counts.}
The synthetic discrete benchmark consists of three scalar sequence-prediction
families: a resonant ARMA process with near-unit-circle oscillatory poles, a
sixth-order IIR cascade, and a nonlinear NARX system with exogenous inputs.
Each discrete task uses $N_{\mathrm{train}}=1024$, $N_{\mathrm{val}}=256$,
$N_{\mathrm{test}}=256$ i.i.d.\ trajectories of length $T=2048$, where these
quantities denote the numbers of training, validation and test trajectories.
The 1D LNO benchmark uses the original $200/50/130$ split of~\cite{cao2024lno}. Five seeds
$\{0,1,2,3,4\}$ are used on all discrete-benchmark tables in the main text and
the appendix; three seeds $\{0,1,2\}$ are used only on the 1D LNO ODE
benchmark.

\paragraph{Compute accounting.}
Neural runs use one NVIDIA A30 GPU (24\,GB) with PyTorch/CUDA; the classical
SysIdentPy baselines run on CPU and contribute less than one hour of wall-clock
time. The supplementary summary CSV files store the mean wall-clock time per
reported entry. Table
\ref{tab:compute_accounting} estimates compute by summing
(\#runs)$\times$(mean training time) over the reported result files used by the
reported tables, figures, and appendix summaries. The final reported runs total
approximately $136$ single-worker wall-clock hours. Re-running the
configuration-selection sweeps that determine the tuned/public settings adds
approximately $21$ hours, for about $157$ single-worker hours in total. Earlier
exploratory screening, discarded protocol variants, and failed/debug runs
required additional compute but are not used to support the paper's claims.

\begin{table}[htbp]
    \centering\footnotesize
    \begin{tabular}{@{}lrrl@{}}
        \toprule
        Experiment family & Runs & Hours & Worker \\
        \midrule
        Main discrete benchmark (matched+tuned) & 90 & 11.2 & GPU \\
        Near-unit-circle difficulty sweep & 75 & 8.1 & GPU \\
        Long-horizon extrapolation, main tasks & 90 & 9.9 & GPU \\
        Long-horizon extrapolation, difficulty sweep & 150 & 16.1 & GPU \\
        Public neural benchmarks & 75 & 2.2 & GPU \\
        Public SysIdentPy baselines & 20 & 0.7 & CPU \\
        Isomorphic $z$/$s$ ablation & 40 & 6.0 & GPU \\
        Training-signal fairness control & 75 & 9.5 & GPU \\
        Internal ZNO ablation & 100 & 39.8 & GPU \\
        IIR stability analysis & 40 & 27.3 & GPU \\
        1D LNO ODE benchmark & 108 & 5.0 & GPU \\
        \midrule
        Reported-results subtotal & 863 & 136.0 & mixed \\
        Configuration-selection sweeps & -- & 21.1 & GPU \\
        \midrule
        Approximate total including selection & -- & 157.1 & mixed \\
        \bottomrule
    \end{tabular}
    \caption{Compute accounting for the submitted experiments. Hours are
    single-worker wall-clock estimates obtained from the supplementary summary CSV
    files and rounded to one decimal place. ``Runs'' counts independent
    training or evaluation jobs in the reported result files; the
    configuration-selection sweeps are summarized separately because they are
    used to choose the tuned/public configurations rather than reported as
    final mean and standard-deviation results.}
    \label{tab:compute_accounting}
\end{table}

\paragraph{S4D tuned configuration.}
The tuned-best S4D configuration of Table~\ref{tab:discrete_main} is chosen from
a state-dimension sweep on the validation split. The validation-selected state
dimensions are
$d_{\mathrm{state}}=16$ on NARX and IIR and $d_{\mathrm{state}}=24$ on resonant
ARMA. The tuned-best S4D numbers of the main table and of the extrapolation
study (Table~\ref{tab:extrapolation}) are confirmed with 5 seeds under this
configuration.

\paragraph{Result files.}
The supplementary material includes the CSV files behind every table and figure,
with per-file documentation of the run commands and data provenance.

\begin{table}[htbp]
    \centering\small
    \begin{tabular}{@{}llrrrrrr@{}}
        \toprule
        Task & Model & $w$ & $L$ & $r$ / $d_{\mathrm{state}}$ / modes & $K$ & $F$ & Params \\
        \midrule
        \multirow{3}{*}{Resonant ARMA} & FNO  & 4  & 4 & modes\,=\,256 & --  & -- & 8\,589 \\
                                        & S4D  & 20 & 4 & $d_{\mathrm{state}}$\,=\,16 & -- & -- & 8\,577 \\
                                        & ZNO  & 20 & 4 & $r$\,=\,8 & 40 & 4 & 8\,657 \\
        \midrule
        \multirow{3}{*}{IIR cascade}    & FNO  & 4  & 4 & modes\,=\,256 & --  & -- & 8\,661 \\
                                        & S4D  & 20 & 4 & $d_{\mathrm{state}}$\,=\,16 & -- & -- & 8\,757 \\
                                        & ZNO  & 20 & 4 & $r$\,=\,8 & 40 & 4 & 8\,837 \\
        \midrule
        \multirow{3}{*}{Nonlinear NARX} & FNO  & 4  & 4 & modes\,=\,256 & --  & -- & 8\,541 \\
                                        & S4D  & 20 & 4 & $d_{\mathrm{state}}$\,=\,16 & -- & -- & 8\,457 \\
                                        & ZNO  & 20 & 4 & $r$\,=\,8 & 40 & 4 & 8\,537 \\
        \bottomrule
    \end{tabular}
    \caption{Truly matched-budget configurations used by
    Table~\ref{tab:discrete_main}. Columns: hidden width $w$, depth $L$, rank /
    state dimension / Fourier modes, ZNO pole count $K$, and FIR order $F$. For ZNO,
    $K$ denotes the number of poles per latent scalar filter; FNO and S4D do not use
    the ZNO-specific $K$ or $F$ hyperparameters. The three models sit in the narrow
    parameter range $8\,457$--$8\,917$ on every task.}
    \label{tab:matched_configs}
\end{table}

\begin{table}[htbp]
    \centering\small
    \begin{tabular}{@{}llrrrrrrr@{}}
        \toprule
        Task & Model & $w$ & $L$ & $r$ / $d_{\mathrm{state}}$ / modes & $K$ & $F$ & LR & Params \\
        \midrule
        \multirow{3}{*}{Resonant ARMA} & FNO  & 8  & 4 & modes\,=\,256 & --  & -- & $2\!\times\!10^{-3}$ & 34\,073 \\
                                        & S4D  & 20 & 4 & $d_{\mathrm{state}}$\,=\,24 & -- & -- & $5\!\times\!10^{-4}$ & 10\,497 \\
                                        & ZNO  & 20 & 4 & $r$\,=\,12 & 64 & 0 & $2\!\times\!10^{-3}$ & 12\,721 \\
        \midrule
        \multirow{3}{*}{IIR cascade}    & FNO  & 8  & 4 & modes\,=\,256 & --  & -- & $2\!\times\!10^{-3}$ & 34\,217 \\
                                        & S4D  & 20 & 4 & $d_{\mathrm{state}}$\,=\,16 & -- & -- & $5\!\times\!10^{-4}$ & 9\,397 \\
                                        & ZNO  & 20 & 4 & $r$\,=\,12 & 80 & 4 & $2\!\times\!10^{-3}$ & 14\,677 \\
        \midrule
        \multirow{3}{*}{Nonlinear NARX} & FNO  & 4  & 4 & modes\,=\,256 & --  & -- & $2\!\times\!10^{-3}$ & 8\,541  \\
                                        & S4D  & 20 & 4 & $d_{\mathrm{state}}$\,=\,16 & -- & -- & $5\!\times\!10^{-4}$ & 9\,097 \\
                                        & ZNO  & 20 & 4 & $r$\,=\,8 & 16 & 4 & $2\!\times\!10^{-3}$ & 7\,001  \\
        \bottomrule
    \end{tabular}
    \caption{Tuned-best configurations used by Table~\ref{tab:discrete_main}. For ZNO,
    $K$ denotes the number of poles per latent scalar filter; FNO and S4D do not use
    the ZNO-specific $K$ or $F$ hyperparameters. S4D
    state dimensions are selected per task by the validation sweep of
    Appendix~\ref{app:protocol}. S4D uses base learning rate
    $5\!\times\!10^{-4}$ and gradient clipping at $1.0$; all other optimizer
    settings match Table~\ref{tab:matched_configs}.}
    \label{tab:tuned_configs}
\end{table}

\section{Long-horizon extrapolation: full numbers}
\label{app:extrap_full}

\begin{figure}[htbp]
    \centering
    \includegraphics[width=\linewidth]{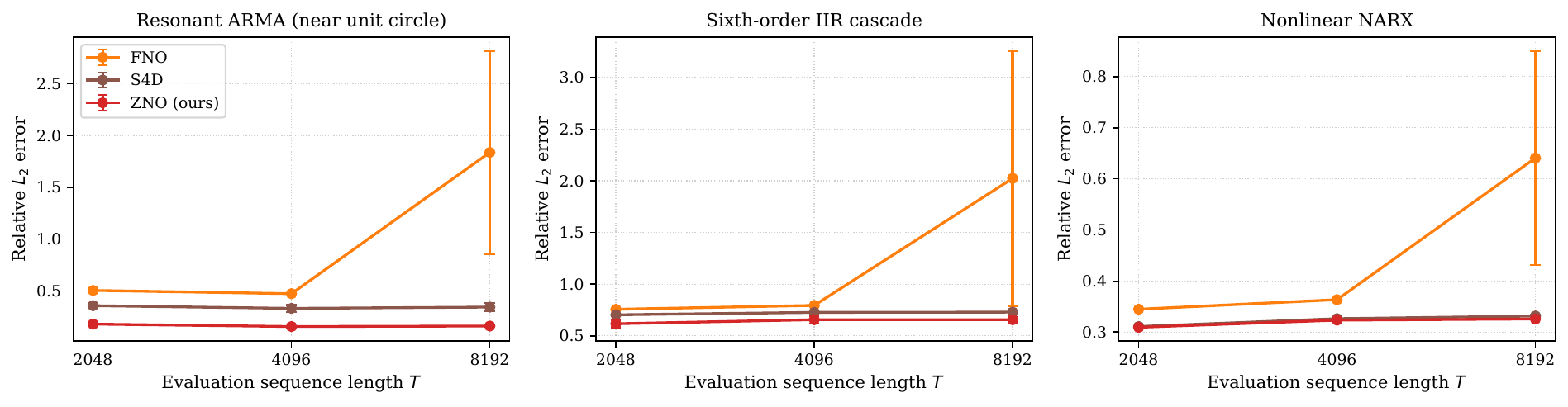}
    \caption{Long-horizon extrapolation at $T\in\{2048,4096,8192\}$ (3 tasks, 5
    seeds). ZNO and S4D are both essentially length-invariant because their layers are causal
    recurrences with length-independent parameter counts. ZNO retains the
    lowest absolute error on every task at every length. FNO error grows sharply
    at $T=8192$ on every task, consistent with its frequency grid being tied to
    the training length.}
    \label{fig:extrapolation}
\end{figure}

\begin{table}[htbp]
    \centering\small
    \begin{tabular}{@{}llrrr@{}}
        \toprule
        Task & Model & $T=2048$ (tuned) & $T=4096$ & $T=8192$\\
        \midrule
        \multirow{3}{*}{Resonant ARMA}    & FNO  & $0.505_{\pm 0.001}$ & $0.474_{\pm 0.001}$ & $1.835_{\pm 0.978}$\\
                                          & S4D  & $0.358_{\pm 0.022}$ & $0.332_{\pm 0.032}$ & $0.345_{\pm 0.038}$\\
                                          & ZNO  & $\mathbf{0.181_{\pm 0.011}}$ & $\mathbf{0.157_{\pm 0.008}}$ & $\mathbf{0.162_{\pm 0.008}}$\\
        \midrule
        \multirow{3}{*}{IIR cascade (6th)}& FNO  & $0.756_{\pm 0.011}$ & $0.794_{\pm 0.006}$ & $2.023_{\pm 1.234}$\\
                                          & S4D  & $0.703_{\pm 0.007}$ & $0.726_{\pm 0.017}$ & $0.729_{\pm 0.014}$\\
                                          & ZNO  & $\mathbf{0.616_{\pm 0.035}}$ & $\mathbf{0.655_{\pm 0.035}}$ & $\mathbf{0.655_{\pm 0.027}}$\\
        \midrule
        \multirow{3}{*}{Nonlinear NARX}   & FNO  & $0.345_{\pm 0.001}$ & $0.363_{\pm 0.001}$ & $0.641_{\pm 0.210}$\\
                                          & S4D  & $0.311_{\pm 0.0004}$& $0.326_{\pm 0.001}$ & $0.331_{\pm 0.003}$\\
                                          & ZNO  & $\mathbf{0.309_{\pm 0.0001}}$& $\mathbf{0.323_{\pm 0.0001}}$& $\mathbf{0.326_{\pm 0.0001}}$\\
        \bottomrule
    \end{tabular}
    \caption{Long-horizon extrapolation (5 seeds, 3 tasks). The same tuned-best
    checkpoints are evaluated at $T\in\{2048,4096,8192\}$. Relative $L_2$ test
    error (mean\,$\pm$\,standard deviation). Both ZNO and S4D are length-stable, with S4D
    ratio $\{0.96,1.04,1.07\}$ at $T=8192$ and ZNO ratio
    $\{0.90,1.06,1.05\}$. ZNO retains the lowest absolute error on every task
    at every length. FNO grows by $3.6\times$, $2.7\times$, $1.9\times$ on
    resonant ARMA, IIR and NARX at $T=8192$.}
    \label{tab:extrapolation}
\end{table}

\paragraph{Extrapolation across the difficulty sweep.}
We rerun the same tuned-best checkpoints of
Section~\ref{sec:experiments_difficulty} at $T\in\{4096,8192\}$ across all
five pole-radius bins. Across all evaluated settings, both ZNO and S4D remain within
$[0.91,1.01]$ of their $T=2048$ error (5-seed means), while FNO inflates on
every bin at $T=8192$ by factors that grow with the memory horizon (bin 0:
$8.1\times$; bin 4: $2.6\times$). ZNO has the lowest mean error on every bin at
every length. The supplementary material includes the numerical summary.

\section{Public nonlinear system-identification benchmarks: protocol and full results}
\label{app:public_protocol}

\paragraph{Benchmarks.}
We evaluate on five public nonlinear system-identification benchmarks:
Silverbox~\cite{wigren2013silverbox} (a Duffing-like electronic circuit),
Wiener--Hammerstein~\cite{schoukens2009wh} (a static nonlinearity sandwiched
between two linear dynamic blocks), cascaded
tanks~\cite{schoukens2016tanks} (a fluid-level benchmark with overflow
nonlinearity), CED~\cite{wigren2017ced} (coupled electric drives), and
EMPS~\cite{janot2019emps} (a linear motor drive).

\paragraph{Protocol.}
All datasets are loaded through the \texttt{nonlinear-benchmarks} Python
package~\cite{beintema2025nonlinearbenchmarks} using the released train/test partitions and converted to scalar
input/output streams before windowing. We take a sequential $80/20$ split on
the training stream to form train/validation partitions; this partition is
shared across all models and seeds. Inputs and outputs are normalized by the
train-stream mean and standard deviation, then windowed into sequences with a
scored segment plus a short initialization prefix. The prefix length is listed
in Table~\ref{tab:public_stats} and is excluded from the reported relative
$L_2$ error, consistent with the public benchmark convention.

\begin{table}[htbp]
    \centering\footnotesize
    \begin{tabular}{@{}lrrrr@{}}
        \toprule
        Benchmark & loaded channels & train / val / test windows & scored $T$ $+$ prefix & sampling period\\
        \midrule
        Silverbox                 & $1/1$ & 391 / 86 / 256 & $2048+50$ & $1.64\!\times\!10^{-3}$\,s\\
        Wiener--Hammerstein       & $1/1$ & 512 / 128 / 256 & $2048+50$ & $1.95\!\times\!10^{-5}$\,s\\
        Cascaded tanks            & $1/1$ & 81 / 4 / 64 & $128+50$ & $4.00\!\times\!10^{0}$\,s\\
        CED                       & $1/1$ & 128 / 16 / 24 & $48+10$ & $2.00\!\times\!10^{-2}$\,s\\
        EMPS                      & $1/1$ & 140 / 23 / 178 & $2048+20$ & $1.00\!\times\!10^{-3}$\,s\\
        \bottomrule
    \end{tabular}
    \caption{Public benchmark statistics. Window counts are after windowing
    each released train/test partition. The scored-$T$ column lists the
    number of time steps included in the reported metric plus the
    initialization prefix supplied to the model but excluded from the metric.
    The loaded-channel column gives input/output channel counts after the
    scalar-stream conversion used by the evaluation code.}
    \label{tab:public_stats}
\end{table}

\paragraph{Configuration selection.}
For each neural architecture we sweep a small grid of architecture knobs and
select the per-benchmark configuration by validation error on the train-stream
$80/20$ split, then run five independent seeds of the selected configuration
on the official test split. The FNO grid varies width $\in\!\{4,6,8\}$ and
Fourier modes $\in\!\{24,32,128,256,384\}$; the S4D grid varies state
dimension $d_{\mathrm{state}}\in\!\{8,12,16,24,32\}$; the ZNO grid varies
pole count $K\in\!\{16,24,32,48,64,80\}$, low-rank factor
$r\in\!\{8,12,16\}$ and FIR order $F\in\!\{0,4,8\}$. Other optimizer settings
match the discrete benchmark (Appendix~\ref{app:protocol}); the ZNO auxiliary
terms are applied only to ZNO. The selected configurations and resulting
parameter counts are listed in Table~\ref{tab:public_selected_tags}.

\paragraph{Classical baseline.}
We include four classical NARMAX baselines implemented in
\texttt{SysIdentPy}~\cite{lacerda2020sysidentpy}: AOLS polynomial, AOLS
Fourier, FROLS polynomial, and FROLS Fourier. Each baseline is swept over a
small grid of (auto/exog) lag, degree and basis order; the candidate with the
best validation error is kept. The classical column of
Table~\ref{tab:public_main} reports the best of the four per benchmark.

\begin{table}[htbp]
    \centering\footnotesize
    \setlength{\tabcolsep}{5pt}
    \begin{tabular}{@{}llllr@{}}
        \toprule
        Benchmark & Model & Selected configuration & Selected arguments & Params\\
        \midrule
        \multirow{3}{*}{Silverbox}            & FNO & w8m384      & \texttt{--width 8 --modes 384}                                & 50\,361\\
                                              & S4D & sd24        & \texttt{--state-dim 24}                                       & 10\,377\\
                                              & ZNO & p80-r12-f8  & \texttt{--num-poles 80 --rank 12 --fir-order 8}               & 14\,569\\
        \midrule
        \multirow{3}{*}{Wiener--Hammerstein}  & FNO & w8m384      & \texttt{--width 8 --modes 384}                                & 50\,361\\
                                              & S4D & sd32        & \texttt{--state-dim 32}                                       & 11\,657\\
                                              & ZNO & p64-r16-f4  & \texttt{--num-poles 64 --rank 16 --fir-order 4}               & 15\,609\\
        \midrule
        \multirow{3}{*}{CED}                  & FNO & w6m24       & \texttt{--width 6 --modes 24}                                 & 2\,443\\
                                              & S4D & sd8         & \texttt{--state-dim 8}                                        & 7\,817\\
                                              & ZNO & p24-r8-f4   & \texttt{--num-poles 24 --rank 8 --fir-order 4}                & 7\,513\\
        \midrule
        \multirow{3}{*}{Cascaded tanks}       & FNO & w6m32       & \texttt{--width 6 --modes 32}                                 & 3\,019\\
                                              & S4D & sd12        & \texttt{--state-dim 12}                                       & 8\,457\\
                                              & ZNO & p32-r12-f4  & \texttt{--num-poles 32 --rank 12 --fir-order 4}               & 9\,769\\
        \midrule
        \multirow{3}{*}{EMPS}                 & FNO & w8m256      & \texttt{--width 8 --modes 256}                                & 33\,977\\
                                              & S4D & sd12        & \texttt{--state-dim 12}                                       & 8\,457\\
                                              & ZNO & p32-r8-f4   & \texttt{--num-poles 32 --rank 8 --fir-order 4}                & 8\,025\\
        \bottomrule
    \end{tabular}
    \caption{Validation-selected configurations behind
    Table~\ref{tab:public_main}. Selection is performed using validation error
    only on the train-stream $80/20$ split; test results are reported from
    five independent seeds of the selected configuration.}
    \label{tab:public_selected_tags}
\end{table}

\paragraph{Validation/test gap.}
Table~\ref{tab:public_val_test_gap} tabulates the best-validation error and
the held-out test error for the selected configuration of every neural model
on every benchmark, together with the test--validation gap. Three
observations: (i) on Silverbox and Wiener--Hammerstein the gap is small for
all three models; (ii) on CED the gap is negative for all three models, i.e.\
the official test split is easier than the held-out tail of the train stream
for this benchmark; (iii) on EMPS the gap is large for FNO ($+0.26$) and
very large for ZNO ($+0.81$), indicating that the selected model does not
transfer across the public train/test regimes for these architectures. We
report this gap rather than re-tuning to absorb it, since re-tuning on test
data would defeat the purpose of an external validity check.

\begin{table}[htbp]
    \centering\footnotesize
    \setlength{\tabcolsep}{6pt}
    \begin{tabular}{@{}llrrr@{}}
        \toprule
        Benchmark & Model & best-val rel.\ $L_2$ & test rel.\ $L_2$ & test $-$ val\\
        \midrule
        \multirow{3}{*}{Silverbox}            & FNO & $0.852$ & $0.984$ & $+0.131$\\
                                              & S4D & $0.043$ & $0.060$ & $+0.018$\\
                                              & ZNO & $0.015$ & $0.021$ & $+0.006$\\
        \midrule
        \multirow{3}{*}{Wiener--Hammerstein}  & FNO & $0.034$ & $0.033$ & $-0.0001$\\
                                              & S4D & $0.030$ & $0.030$ & $+0.001$\\
                                              & ZNO & $0.0027$ & $0.0028$ & $+0.0001$\\
        \midrule
        \multirow{3}{*}{CED}                  & FNO & $0.440$ & $0.204$ & $-0.236$\\
                                              & S4D & $0.362$ & $0.177$ & $-0.185$\\
                                              & ZNO & $0.312$ & $0.150$ & $-0.162$\\
        \midrule
        \multirow{3}{*}{Cascaded tanks}       & FNO & $0.251$ & $0.382$ & $+0.131$\\
                                              & S4D & $0.282$ & $0.324$ & $+0.043$\\
                                              & ZNO & $0.283$ & $0.370$ & $+0.088$\\
        \midrule
        \multirow{3}{*}{EMPS}                 & FNO & $0.057$ & $0.321$ & $+0.264$\\
                                              & S4D & $0.245$ & $0.338$ & $+0.093$\\
                                              & ZNO & $0.227$ & $1.041$ & $+0.815$\\
        \bottomrule
    \end{tabular}
    \caption{Validation/test gap on the public benchmarks for the selected
    configuration of each neural architecture (mean over five seeds).
    Best-val column: relative $L_2$ on the train-stream $80/20$ validation
    split that is used for configuration selection. Test column: relative
    $L_2$ on the official held-out test set. Negative gaps on CED indicate
    that the official test split is easier than the validation tail of the
    train stream; the large positive gap on EMPS for FNO and ZNO indicates
    that the selected model does not transfer between the two regimes.}
    \label{tab:public_val_test_gap}
\end{table}

\section{Isomorphic \texorpdfstring{$z$-plane versus $s$-plane}{z vs s} ablation}
\label{app:isomorphic}

The ZNO layer combines several modeling choices (low-rank MIMO factorization,
smooth stability reparameterization, and a discrete recurrent scan) with a fused
implementation of the scan and the specific $z$-plane pole form. To isolate the
pole-coordinate choice, we replace only the pole parameterization inside an
otherwise identical ZNO layer. For
each latent channel and pole index, the $z$-plane variant stores
$(\tilde\rho,\phi)\in\R^2$ and produces
$p=\rho_{\max}\operatorname{sigmoid}(\tilde\rho)\e^{\ii\phi}$; the isomorphic $s$-plane
variant stores $(\tilde\alpha,\omega)$, maps them to a stable
$\mu=-\mathrm{softplus}(\tilde\alpha)+\ii\omega$, then to a discrete
pole via $p=\e^{\mu \Delta t}$ with $\Delta t=1$. All other layer components and
training details (low-rank projections, residues, FIR branch, recurrent scan,
pole-safety regularizer, optimizer, suffix loss) are shared, as is the fused
implementation. The anchor
configuration is $r=8$, $F=4$; $K=32$ for ARMA, $K=48$ for IIR ($8\,145$ /
$9\,349$ params); the tuned-best row uses the ZNO tuned-best configurations
from Appendix~\ref{app:protocol}.

\begin{table}[htbp]
    \centering\small
    \begin{tabular}{@{}llrrr@{}}
        \toprule
        Task & Configuration & $z$-plane (ZNO) & $s$-plane isomorphic & $\Delta(s-z)$\\
        \midrule
        Resonant ARMA     & matched-budget & $0.219_{\pm 0.009}$ & $0.220_{\pm 0.011}$ & $+0.0006$ \\
        Resonant ARMA     & tuned-best     & $0.181_{\pm 0.011}$ & $0.179_{\pm 0.010}$ & $-0.0020$ \\
        IIR cascade (6th) & matched-budget & $0.661_{\pm 0.008}$ & $0.665_{\pm 0.023}$ & $+0.0039$ \\
        IIR cascade (6th) & tuned-best     & $0.616_{\pm 0.035}$ & $0.623_{\pm 0.036}$ & $+0.0068$ \\
        \bottomrule
    \end{tabular}
    \caption{Isomorphic $z$-plane vs $s$-plane ablation (5 seeds). On every row
    the mean gap $\Delta(s-z)$ is small relative to the seed-level standard deviation of
    either variant, so no practically meaningful difference is observed at
    this sample size. A paired-seed test at larger $n$ would be needed for a
    stronger equivalence claim.}
    \label{tab:isomorphic_splane}
\end{table}

The two variants are not separated at this sample size: the largest mean gap
is $0.0068$ on the tuned-best IIR cascade, well within the 5-seed standard deviations of
$\sim\!0.035$, and every row has $|\Delta(s-z)|$ below either variant's standard deviation.
Once the low-rank rational recurrent layer is fixed, the choice of a $z$-plane
rather than an isomorphic $s$-plane pole parameterization does not appear to be an
independent source of the accuracy gain over FNO and S4D at the scale of our
experiments. The $z$-domain rational formulation remains central to the model design:
it removes the free hyperparameter $\Delta t$, places the stability boundary at the
unit circle (single smooth sigmoid bound), and gives a direct interpretability
read-out (Appendix~\ref{app:traces}).

\section{1D LNO ODE benchmark}
\label{app:1d_benchmark}

We rerun the LNO benchmark of~\cite{cao2024lno} (pendulum, Duffing, Lorenz,
two damping scenarios each) using the original data generators and splits,
and compare ZNO against LNO and FNO under two protocols. This benchmark family
consists of data generated from continuous-time ODE systems, rather than native
discrete-time system-identification tasks.
\begin{itemize}
    \item \textbf{Matched-budget.} All three models at $\sim\!7$--$9$k
    parameters, with identical optimizer and schedule.
    \item \textbf{Tuned-best.} Each model in its reported tuned configuration
    from~\cite{cao2024lno}; for FNO this is a $\sim\!2$M-parameter model with
    256 Fourier modes.
\end{itemize}

\begin{figure}[htbp]
    \centering
    \includegraphics[width=\linewidth]{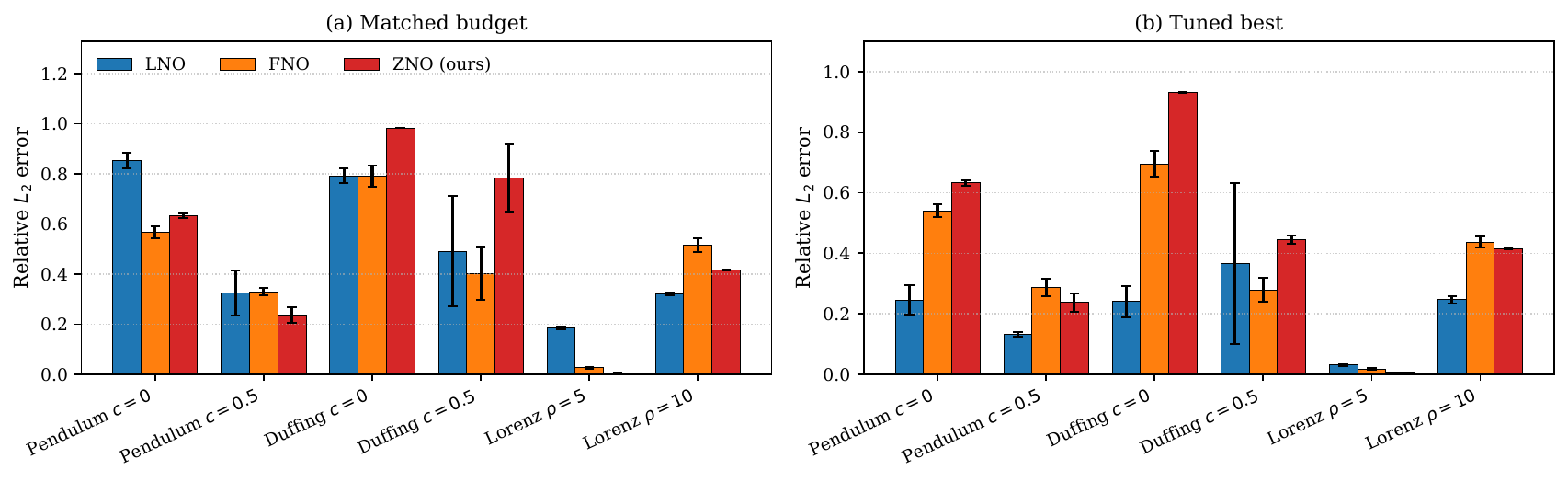}
    \caption{1D LNO ODE benchmark, relative $L_2$ test error
    (mean\,$\pm$\,standard deviation, 3 seeds). (a) Matched-budget protocol ($\sim\!7$--$9$k
    params). (b) Tuned-best protocol. ZNO has the lowest error on Lorenz\,$\rho=5$
    and is competitive elsewhere; LNO and FNO each have the lowest error on some cases, as expected given
    that the benchmark was designed around continuous-time ODEs.}
    \label{fig:benchmark_1d}
\end{figure}

\paragraph{Results.}
Under matched budget, ZNO has the lowest error on Lorenz\,$\rho=5$ ($\relL=0.007$ vs $0.18$
LNO, $0.026$ FNO) and Pendulum\,$c=0.5$; LNO has the lowest error on Lorenz\,$\rho=10$; FNO
has the lowest error on the two Duffing cases and on Pendulum\,$c=0$. Under tuned-best LNO
takes the lead on most cases but only with its best small-model
configuration, and ZNO retains the lowest error on Lorenz\,$\rho=5$. The
benchmark was designed for continuous-time ODEs and LNO/FNO have a natural
fit to it. We include it to show that ZNO remains stable and competitive on
data generated from continuous-time ODE benchmarks and to provide a like-for-like comparison
on the original training pipeline.

\section{Reference LNO on discrete benchmarks}
\label{app:lno_pilot}

We ran the reference LNO layer on an exploratory discrete benchmark covering
ARMA, DSS, IIR and NARX families. Training diverged on every case: the relative
$L_2$ test error settled in the range $2.6\times 10^{10}$ to
$1.3\times 10^{11}$ across seeds. We therefore report this result as a negative
check and do not include the reference LNO in the main discrete tables. The
separate $z$/$s$ ablation in Appendix~\ref{app:isomorphic} keeps the ZNO
architecture fixed and isolates only the pole-coordinate parameterization; it is
not intended as an LNO baseline.

\section{Training-signal fairness control}
\label{app:suffix_fairness}

The ZNO objective includes a suffix relative-$L_2$ term
($\lambda_{\mathrm{suf}}=10^{-2}$) and the pole-safety
regularizer~\eqref{eq:safety}. These auxiliary terms are not used by FNO and
S4D in the main training protocol.
To verify that the ZNO advantage in Table~\ref{tab:discrete_main} is not an
artifact of this training-signal asymmetry, we ran at the matched parameter
    budget a symmetric control with five seeds for each model-task setting:
    (i) FNO and S4D retrained
with the same suffix loss (\emph{suffix-matched}); (ii) ZNO retrained
with the suffix term turned off (\emph{suffix-off}) and with both the suffix term
and the pole-safety regularizer turned off (\emph{suffix/pole-off}). All
other hyperparameters are identical to Table~\ref{tab:matched_configs}.
Suffix-loss matching FNO and S4D leaves their test error essentially unchanged
($\Delta<0.001$ everywhere). Removing both auxiliary terms from ZNO leaves
its error within one seed-level standard deviation of baseline on every task. The fairness
control therefore does not erase the ZNO advantage on resonant ARMA or NARX;
on IIR the baseline ordering favors S4D under both the main and the
control protocol.

\begin{table}[htbp]
    \centering\footnotesize
    \setlength{\tabcolsep}{2.5pt}
    \begin{tabular}{@{}lrrrrr@{}}
        \toprule
        Task & FNO + suffix & S4D + suffix & ZNO base & ZNO no suffix & ZNO no suffix/pole\\
        \midrule
        Resonant ARMA  & $0.515_{\pm 0.005}$ & $0.387_{\pm 0.011}$ & $0.203_{\pm 0.017}$ & $0.199_{\pm 0.013}$ & $0.198_{\pm 0.013}$\\
        IIR cascade    & $0.851_{\pm 0.012}$ & $0.704_{\pm 0.011}$ & $0.727_{\pm 0.136}$ & $0.736_{\pm 0.132}$ & $0.737_{\pm 0.132}$\\
        Nonlinear NARX & $0.345_{\pm 0.001}$ & $0.311_{\pm 0.0001}$& $0.309_{\pm 0.0001}$& $0.309_{\pm 0.0001}$& $0.309_{\pm 0.0001}$\\
        \bottomrule
    \end{tabular}
    \caption{Training-signal fairness control (5 seeds). All entries use the
    matched parameter budget of Table~\ref{tab:discrete_main}. Suffix-loss matching
    FNO and S4D or removing the ZNO-only auxiliary terms does not change the
    model ordering: ZNO has the lowest mean error on resonant ARMA and NARX,
    S4D has the lowest mean error on IIR, and FNO has the highest mean error on
    every task.}
    \label{tab:suffix_fairness}
\end{table}

\section{Internal ZNO ablation}
\label{app:ablation}

We ablate the four internal design knobs: number of poles per layer ($K$),
low-rank factor ($r$), FIR order ($F$) and safe radius ($\rho_{\mathrm{safe}}$).
The ablation is performed on the resonant ARMA and sixth-order IIR cascade
tasks, with the ZNO anchor configuration $r=8$, $F=4$,
$\rho_{\mathrm{safe}}=0.95$; $K=32$ for ARMA and $K=48$ for IIR ($8\,145$ /
$9\,349$ params). This anchor is shared with the isomorphic $s$-plane
ablation (Appendix~\ref{app:isomorphic}) so the two share a common reference
point. Five seeds per configuration.

\begin{figure}[htbp]
    \centering
    \includegraphics[width=\linewidth]{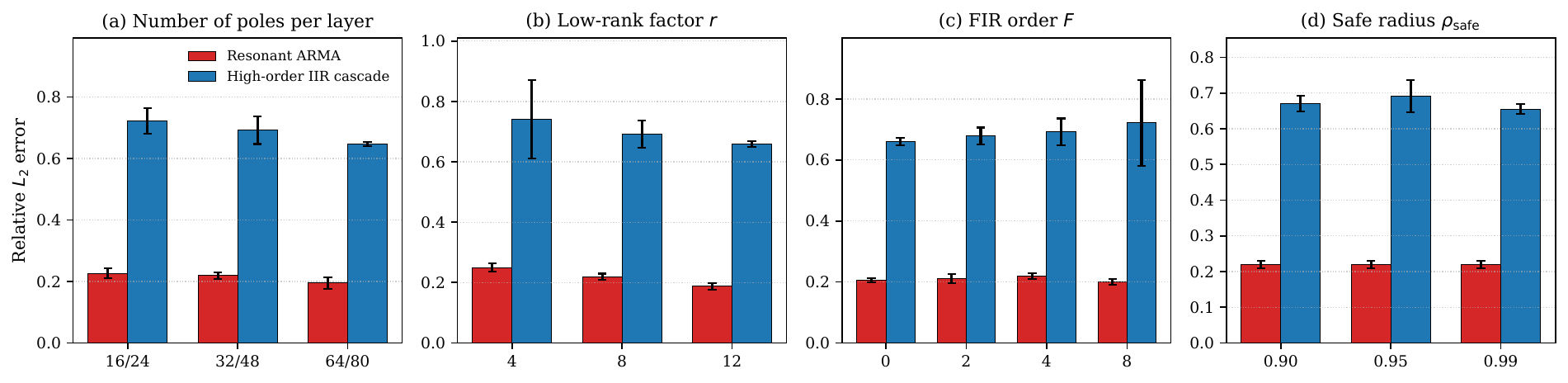}
    \caption{Internal ablation of ZNO on resonant ARMA and sixth-order IIR
    cascade (5 seeds). (a) Doubling $K$ lowers error on both tasks. (b)
    Raising $r$ from $4$ to $12$ yields a clear improvement. The $r=4$ setting is
    under-parameterized and diverges on IIR ($0.741_{\pm 0.130}$), $r=12$
    approaches tuned-best. (c) FIR order is least sensitive: $F=0$ matches
    baseline on ARMA; $F=8$ increases variance and yields the least stable IIR
    result among the FIR settings. (d) Safe radius $\rho_{\mathrm{safe}}\in\{0.90,0.95,0.99\}$
    has almost no effect, confirming the pole-safety regularizer acts as a
    soft constraint rather than a driver of the result.}
    \label{fig:ablation}
\end{figure}

\begin{table}[htbp]
    \centering\small
    \begin{tabular}{@{}lrr@{}}
        \toprule
        Configuration & Resonant ARMA & IIR cascade (6th)\\
        \midrule
        baseline ($r=8,F=4,K=\{32,48\},\rho_{\mathrm{safe}}=0.95$)
          & $0.219_{\pm 0.011}$ & $0.692_{\pm 0.044}$\\
        rank $r=4$                              & $0.250_{\pm 0.013}$ & $0.741_{\pm 0.130}$\\
        rank $r=12$                             & $\mathbf{0.188_{\pm 0.011}}$ & $0.659_{\pm 0.010}$\\
        poles $K=\{16,24\}$                     & $0.227_{\pm 0.016}$ & $0.721_{\pm 0.041}$\\
        poles $K=\{64,80\}$                     & $0.195_{\pm 0.019}$ & $\mathbf{0.647_{\pm 0.007}}$\\
        FIR order $F=0$                         & $0.205_{\pm 0.007}$ & $0.660_{\pm 0.012}$\\
        FIR order $F=2$                         & $0.211_{\pm 0.015}$ & $0.678_{\pm 0.028}$\\
        FIR order $F=8$                         & $0.200_{\pm 0.010}$ & $0.722_{\pm 0.140}$\\
        safe radius $\rho_{\mathrm{safe}}=0.90$ & $0.219_{\pm 0.011}$ & $0.671_{\pm 0.023}$\\
        safe radius $\rho_{\mathrm{safe}}=0.99$ & $0.219_{\pm 0.011}$ & $0.656_{\pm 0.013}$\\
        \bottomrule
    \end{tabular}
    \caption{Internal ablation (5 seeds). Relative $L_2$ test error on the
    resonant ARMA and sixth-order IIR cascade tasks with one knob perturbed
    at a time away from the ZNO anchor. Bold indicates the lowest error per
    column. The low-rank
    factor and the pole count are the dominant knobs; FIR order and safe
    radius are comparatively insensitive.}
    \label{tab:ablation}
\end{table}

\section{IIR matched-budget stability analysis}
\label{app:iir_stability}

Under the matched-budget protocol, the sixth-order IIR ZNO run has one divergent
seed that inflates the 5-seed standard deviation to $0.138$
(Section~\ref{sec:experiments_discrete}). The isomorphic $s$-plane variant
shows the same symptom, hinting at an optimization issue of the low-rank
rational layer rather than a $z$-plane-specific pathology. To verify this we
swept the optimizer configuration on the same matched budget with 5 seeds per
optimizer setting. Stronger gradient clipping (clip $=0.5$) with a slightly larger base
learning rate (\texttt{lr=1e-3}) converts the divergent run into a stable
one: the 5-seed mean moves from $0.724$ to $0.706$ and the standard deviation from $0.138$
to $0.023$. The same recipe stabilizes the $s$-plane isomorphic variant to
$0.693_{\pm 0.014}$, confirming the fix is parameterization agnostic. We do
not promote this separate optimizer recipe into the main table because the
main protocol uses a single shared optimizer across tasks.

\begin{table}[htbp]
    \centering\small
    \begin{tabular}{@{}llr@{}}
        \toprule
        Model & Variant & IIR cascade (6th) rel.\ $L_2$ (5 seeds)\\
        \midrule
        ZNO (main Table~\ref{tab:discrete_main}) & baseline & $0.724_{\pm 0.138}$\\
        ZNO                                      & lr1e-3, clip 0.25               & $0.744_{\pm 0.033}$\\
        ZNO                                      & lr1e-3, clip 0.5                & $\mathbf{0.706_{\pm 0.023}}$\\
        ZNO                                      & lr8e-4, clip 0.25, safe 0.90    & $0.742_{\pm 0.042}$\\
        ZNO                                      & lr8e-4, clip 0.5                & $0.725_{\pm 0.032}$\\
        \midrule
        ZNO $s$-plane iso & lr1e-3, clip 0.25               & $0.749_{\pm 0.049}$\\
        ZNO $s$-plane iso & lr1e-3, clip 0.5                & $\mathbf{0.693_{\pm 0.014}}$\\
        ZNO $s$-plane iso & lr8e-4, clip 0.25, safe 0.90    & $0.743_{\pm 0.038}$\\
        ZNO $s$-plane iso & lr8e-4, clip 0.5                & $0.730_{\pm 0.041}$\\
        \bottomrule
    \end{tabular}
    \caption{Sixth-order IIR cascade stability analysis at the matched budget
    (5 seeds). Stronger gradient clipping at \texttt{lr=1e-3} moves the ZNO
    5-seed standard deviation from $0.138$ to $0.023$ and the mean from $0.724$ to $0.706$;
    the same recipe produces $0.693_{\pm 0.014}$ on the $s$-plane isomorphic
    variant. The divergence is an optimization artifact of the low-rank
    rational layer, not a $z$-plane-specific issue.}
    \label{tab:iir_stability}
\end{table}

\section{Accuracy vs training-time trade-off}
\label{app:tradeoff}

This figure reports a practical wall-clock comparison under the tuned-best
protocol. It is not intended as a FLOP-level fairness claim because ZNO uses a
fused Triton scan whereas FNO and S4D use their reference PyTorch
implementations.

\begin{figure}[H]
    \centering
    \includegraphics[width=0.65\linewidth]{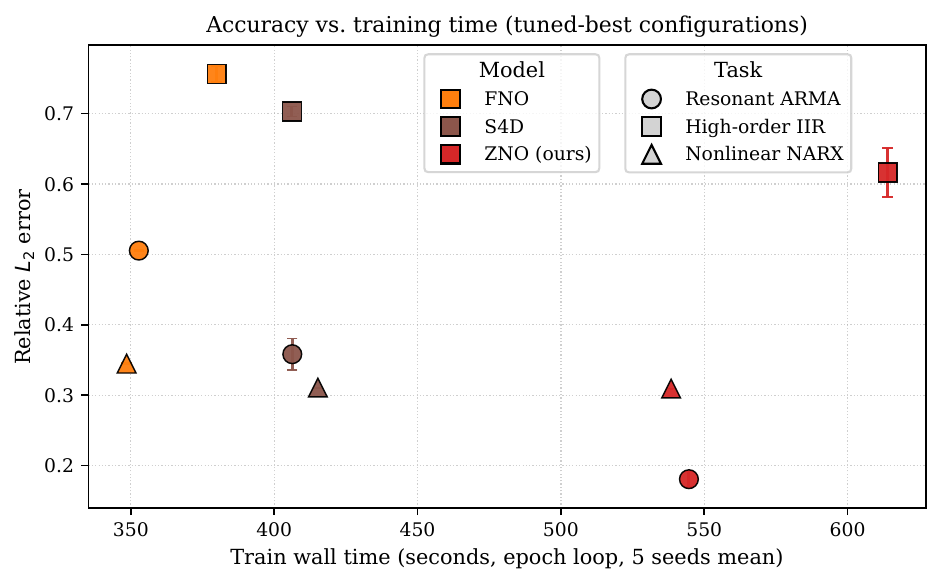}
    \caption{Accuracy vs training time under the tuned-best protocol (5-seed
    mean; error bars: standard deviation). ZNO requires more wall-clock time for lower test error.
    FNO and S4D use plain PyTorch implementations; ZNO uses the fused Triton
    scan, so this is a practical cost comparison rather than a matched-kernel
    comparison.}
    \label{fig:tradeoff}
\end{figure}

\section{Learned \texorpdfstring{$z$}{z}-plane pole maps and qualitative traces}
\label{app:traces}

This diagnostic supports the interpretability claim by plotting the learned
poles $p$ in the discrete recurrence~\eqref{eq:state_recurrence_layer}.
Because ZNO parameterizes $p$ directly, the poles are readable on the unit disk:
resonant ARMA places poles near the unit circle at
$\phi\!\in\!(0.05\pi,0.45\pi)$, the IIR task pushes later-layer poles toward
$|p|\!\approx\!0.95$, and NARX concentrates poles near the real axis at
$|p|\!\approx\!0.4$--$0.7$. LNO and S4D use continuous-time parameters, and FNO
uses fixed Fourier modes, so this unit-disk read-out is not as direct for those
baselines; Appendix~\ref{app:isomorphic} separately isolates the coordinate
parameterization.

\begin{figure}[htbp]
    \centering
    \includegraphics[width=0.9\linewidth]{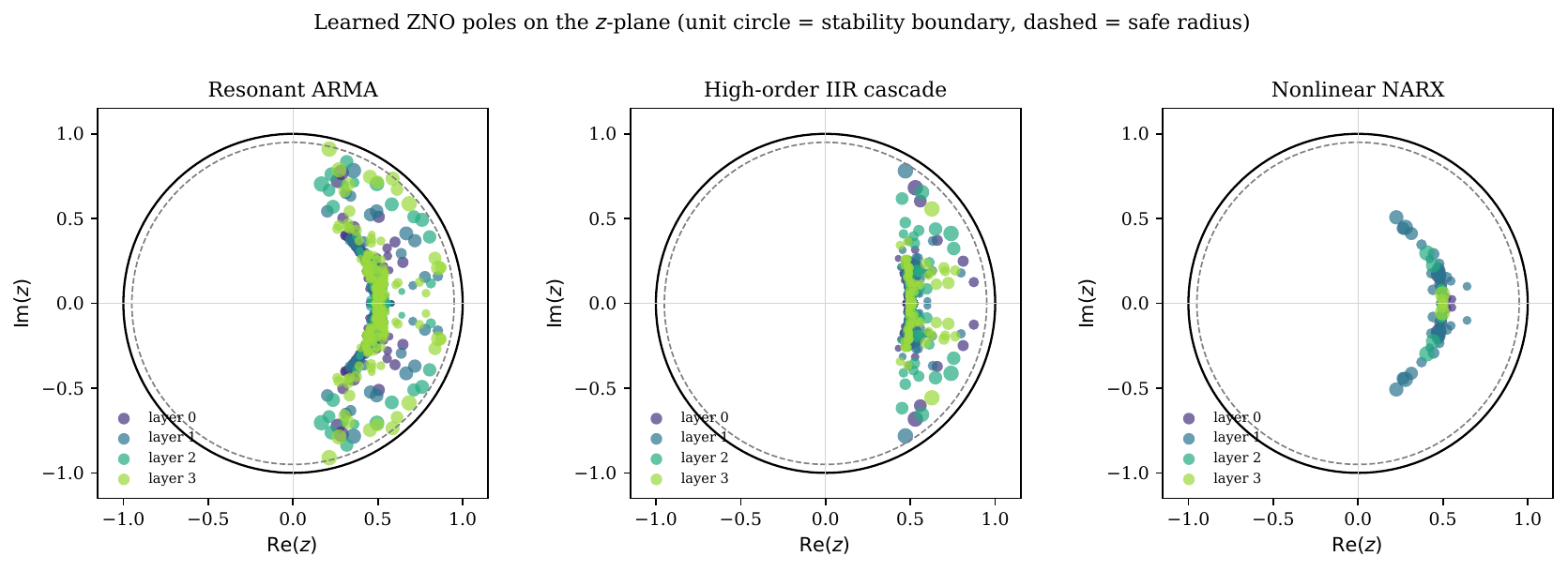}
    \caption{Learned ZNO poles on the $z$-plane per task. Color: layer depth;
    marker size: residue magnitude. Solid/dashed circles: stability boundary
    and safe radius $\rho_{\mathrm{safe}}=0.95$. The maps show near-unit-circle
    oscillatory poles for resonant ARMA, later-layer IIR poles near the safe
    radius, and NARX poles near the real axis.}
    \label{fig:zplane_poles}
\end{figure}

\begin{figure}[htbp]
    \centering
    \includegraphics[width=\linewidth]{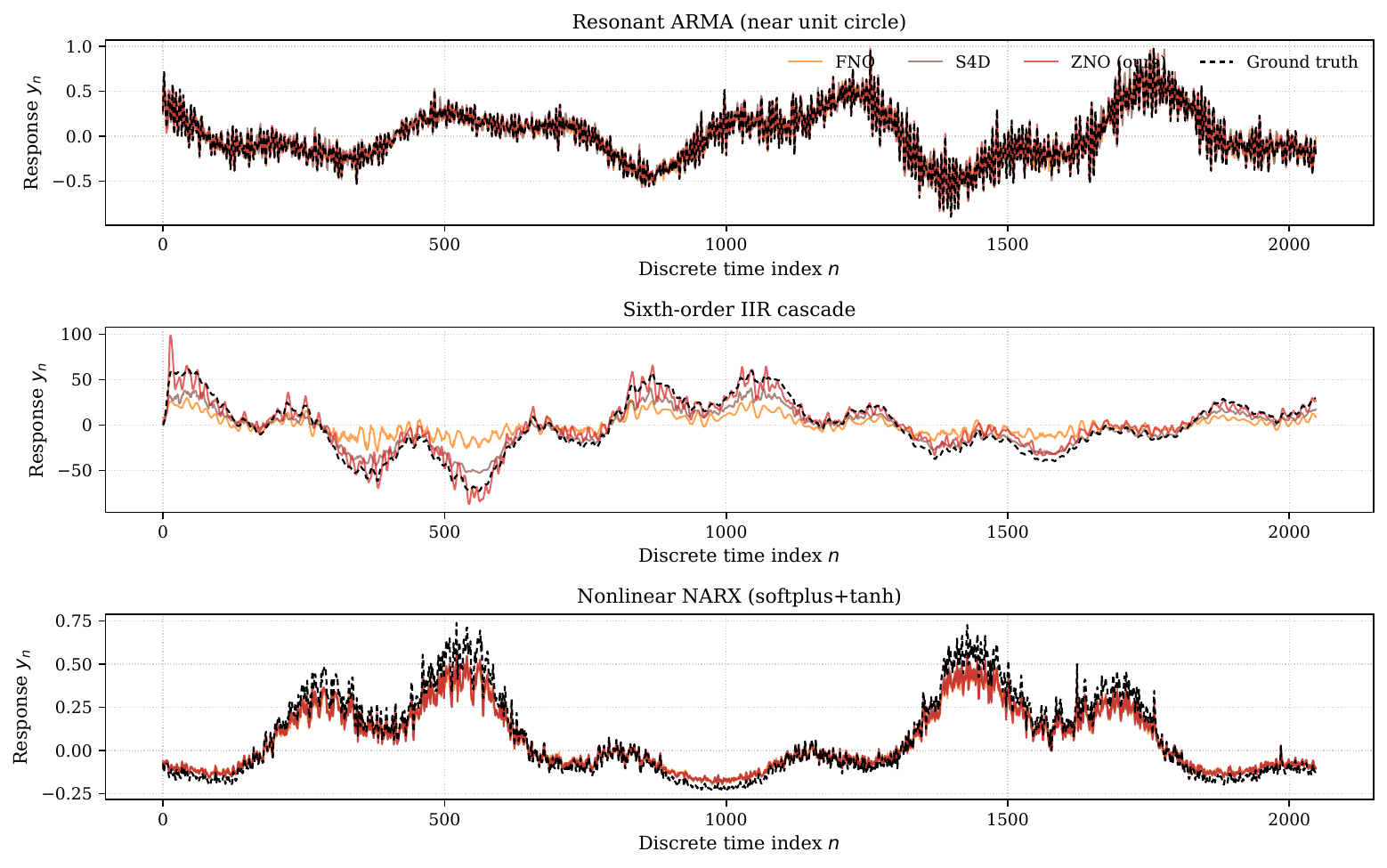}
    \caption{Qualitative comparison on one held-out test trajectory per task.
    Black dashed: ground truth. Colored: FNO (orange), S4D (brown), ZNO (red). On the
    resonant ARMA panel, ZNO has the smallest trajectory error and most closely
    overlays the ground truth. On the sixth-order IIR cascade, FNO substantially
    underestimates the response amplitude, while S4D and ZNO are closer to the
    ground truth on this trajectory. On NARX, all three models follow the main
    low-frequency response, with S4D and ZNO nearly overlapping.}
    \label{fig:prediction_traces}
\end{figure}

\end{document}